\documentclass[review]{elsarticle}

\usepackage{lineno,hyperref}
\modulolinenumbers[5]

\pdfoutput=1

\usepackage{amsmath,amssymb,amsfonts}
\usepackage{algorithmic} 
\usepackage{graphicx}
\usepackage{textcomp}
\usepackage{xcolor}
\usepackage{subcaption}

\usepackage{subfiles} 
\usepackage[ruled, lined, linesnumbered, commentsnumbered, longend]{algorithm2e}
\usepackage{paralist}
\usepackage{afterpage}
\usepackage{placeins}

\newcommand{\ie}{\textit{i}.\textit{e}., }
\newcommand{\eg}{\textit{e}.\textit{g}. }

\newcommand{\beastEach}{\emph{HSel}}
\newcommand{\bestPooled}{\emph{PSel}}
\newcommand{\ParetoSurvival}{\emph{PSv}}
\newcommand{\newSurvival}{\emph{ISv}}

\journal{Journal of Knowledge-Based Systems}

\begin{document}

\begin{frontmatter}

\title{Automated Circuit Sizing with Multi-objective Optimization based on Differential Evolution and Bayesian Inference}





\author[upb]{Cătălin Vișan}
\author[upb]{Octavian Pascu}
\author[upb]{Marius Stănescu}
\author[upb,infineon]{Elena-Diana Șandru}
\author[infineon]{Cristian Diaconu}
\author[infineon]{Andi Buzo}
\author[infineon]{Georg Pelz}
\author[upb]{Horia Cucu}

\address[upb]{University "Politehnica" of Bucharest}
\address[infineon]{Infineon Technologies}

\begin{abstract}
    With the ever increasing complexity of specifications, 
    manual sizing for analog circuits recently became very challenging.
    Especially for innovative, large-scale circuits designs, with tens of design variables, 
    operating conditions and
    conflicting objectives to be optimized,
    design engineers spend many weeks, running time-consuming simulations,
    in their attempt at finding the right configuration.
    Recent years brought machine learning and optimization techniques to the field of analog circuits design,
    with evolutionary algorithms and bayesian models showing good results for circuit sizing.
    
    In this context, we introduce a design optimization method based on Generalized Differential Evolution 3 (GDE3)
    and Gaussian Processes (GPs).
    The proposed method is able to perform sizing for complex circuits with a large number of design variables 
    and many conflicting objectives to be optimized.
    While state-of-the-art methods reduce multi-objective problems to single-objective optimization 
    and potentially induce a priori bias, 
    we search directly over the multi-objective space using Pareto dominance 
    and ensure that \emph{diverse solutions} are provided to the designers to choose from.
    To the best of our knowledge, the proposed method is the first to specifically address the diversity of the solutions, while also focusing on minimizing the number of simulations required to reach feasible configurations.
    We evaluate the introduced method on two voltage regulators showing different levels of complexity
    and we highlight that the proposed innovative candidate selection method and survival policy leads
     to obtaining feasible solutions, with a high degree of diversity, much faster than with GDE3 or Bayesian Optimization-based 
     algorithms.
\end{abstract}

\begin{keyword}
\texttt{Evolutionary Algorithms \sep Gaussian Processes \sep Circuit Design \sep Multi-objective Optimization}
\end{keyword}

\end{frontmatter}




\section{Introduction}\label{sec:intro}


Manual design for analog circuits is currently performed by highly-qualified, experienced
engineers using computer-aided design (CAD) tools. However, with the continuous increase
of specifications and requirements complexity, even the top engineers face difficulties. 
Taking this into
consideration, circuit design automation based on Machine Learning (ML) is getting more
and more attention over the past years. ML-based CAD tools could enable engineers to
produce \textit{more efficient} solutions \textit{faster}, and the learning capabilities
of the algorithms could help designers shed more light on the behavior and the dependency
of circuit performance on influential factors. As a result, this could reduce R\&D costs
and accelerate the time to market for various products. Typically, there are two major
components regarding circuit design automation, \ie topology selection and device
sizing. In this paper, we focus on device sizing for analogue/mixed signal circuits and
apply the proposed method to a voltage regulator with $27$ design variables and $11$ responses (out
of which three are target objectives), all $11$ with specification constraints, as well as on a slightly complex
2.5V regulator with $8$ design variables and $6$ responses, all of them being considered as target objectives. Most
research in this area tests on circuits with fewer than $10$ design variables, and even
though very recent methods \cite{he2020} use problems with $24$ and $20$ design
variables, they only have one target objective and six and two constraints respectively.



There are two major conventional approaches for device sizing: model-based and
simulation-based. In the first approach, designers encode their knowledge into
a mathematical model that models the circuit's outputs as functions of the design
parameters. With such a model at hand, circuit sizing becomes an optimization problem.
Regression is also a viable approach, when mathematical expressions cannot be derived
easily. With this respect, one of the most popular methods for modeling circuits is the
one based on geometric programming \cite{boyd2005,Sayed2019}: circuit responses are expressed as
posynomial and monomial functions of design parameters. There are a few advantages
of using this method: (i) low computational cost for model evaluation, (ii) the model is
reusable for further optimization, (iii) posynomial models guarantee finding the global
optimum. However, this method has a significant drawback: there is an intrinsic deviation
between the model and the real circuit. In consequence, this usually leads to finding --
based on the model -- optima that can be very different from the real circuit optima.

In the simulation-based approach, the circuit performance is considered as a black-box
function and the optimization of design parameters is based solely on circuit
simulations. There are many global optimization algorithms, 
such as Simulated Annealing (SA) \cite{848091}
\cite{Panerati2017}, Particle Swarm Intelligence (PSO) \cite{MISHRA2014111},
and almost all evolutionary algorithms that can be used for such optimizations
\cite{LIU2009137} \cite{RePEc:eee:energy:v:125:y:2017:i:c:p:681-704}. One of the main
advantages of this method is that the aforementioned optimization algorithms usually find the
global optima. The main drawback of this approach resides in the fact that convergence
requires many simulations, which can lead to a very long optimization time.


To mitigate the disadvantages of both approaches, methods based on surrogate models
have been proposed and had a massive success in this area \cite{Lyu2018} \cite{8116661}.
This hybrid methodology is introducing an online evolving model, which
is updated incrementally through the optimization procedure. Due to the lack of data at
the beginning of the optimization, the model can be built by random sampling data or
through the Design of Experiments method (DoE). 
At each optimization iteration, more promising points are explored, and used to update the model.
One of the first algorithms based on this approach
is GASPAD \cite{Liu2014}, which was proposed to synthesize microwave circuits.

The Gaussian Processes (GPs) are the most widely used surrogate models, thanks to the fact
that besides the predicted values, they also provide a measure of the prediction
uncertainty. Moreover, the Gaussian structure used makes learning efficient and GPs
usually work well with little available data. The main drawback is the high computational
cost of training and prediction, up to \emph{O(N\textsuperscript{3})}.

Lately, Neural Networks (NNs) have achieved comparable performances to the GPs, with a
reduced computational cost of \emph{O(N)} for training and constant prediction time
\cite{8714788}. Compared to just GPs regression models, with explicitly
defined kernel functions, neural-network-based models provide, in theory, more accurate
predictions and thus should accelerate the follow-up optimization procedure. However,
they usually require more data to obtain similar generalization power without
overfitting, compared to GP models.

Approaches based on NNs, more specifically Deep Reinforcement Learning (DRL), 
have started to be used in circuit design. The paper \cite{wang2018learning} presents a single-objective DRL algorithm, which 
is developed further in \cite{wang2020gcn}, that has better performance than popular single-objective 
algorithms such as Bayesian Optimization (BO) and Multi-objective Acquisition Ensemble (MACE).
In order to apply the DRL algorithm, the circuit designer must create an objective function 
they named "Figure of Merit" which combines the effect of the circuit responses.


Most surrogate-model based approaches are tested on single-objective problems
\cite{Liu2014,Lyu2018-mb} or transform  the multi-objective problem into a
single-objective one via scalarization \cite{Lyu2018} \cite{8116661}.
In this work, we focus on the multi-objective aspect of the problems and the associated challenges, 
in particular convergence speed and diversity management. 

The main contributions of this paper are summarized as follows. Firstly, we propose a
multi-objective optimization method based on combining a Generalized Differential
Evolution 3 \cite{Kukkonen2005-mj} inspired algorithm with GPs that is designed to work well on
problems with a high number of variables and many objectives and constraint
specifications. Secondly, we propose innovative population survival policies and
candidate selection algorithms that reduce the number of real simulations required to
finish the optimization process and \emph{preserve the diversity of the solutions}.
To the best of our knowledge, the proposed method represents the first one to specifically address solution diversity 
by performing the search directly over the multi-objective space, using Pareto dominance.
Finally, we evaluate the proposed method with state-of-the-art evolutionary algorithms and Bayesian-optimization algorithms on two real voltage regulators.




\section{Background}\label{sec:background}

The analog circuit sizing problem can be formulated as a constrained optimization
problem: minimize $f(x)$ such that $c_{i}(x) < 0$, $i \in 1 ... N_{c}$, where $x \in
R\textsuperscript{d}$ represents the d design variables, $f(x)$ represents the figure of merit
of the circuit, and $c_{i}(x)$ is the i-th constraint of the $N_{c}$ constraint
functions that need to be fulfilled. This formulation only stands for Single-objective
optimization, but if we turn the attention to Multi-Objective Optimization, the
formulation written above should be extended as follows: minimize $f_{1}(x), ...,
f_{m}(x)$ such that $c_{i}(x) < 0$, $i \in 1 ... N_{c}$.

In a single-objective optimization problem, the best design can be found with a minimum
value of the target objective. However, in multi-objective optimization minimizing one
objective is usually in conflict with minimizing another one, and there rarely is a
global optimal solution. Instead, a set of non-dominated solutions which approximated the
Pareto Front (PF) is provided. 

In this section, we start introducing notions about the constraint handling (subsection \ref{subsec:metrics}), 
followed by the state-of-the-art multi-objective performance metrics (subsection \ref{subsec:HV}). Then, we present 
the most promising classes of algorithms employed for circuit sizing assignment found in the literature:
Evolutionary Algorithms (subsection \ref{subsec:EAs}) and Surrogate Models (subsection \ref{subsec:surrmodels}).

\subsection{Constraint Handling}\label{subsec:metrics}

Constraint handling is one of the most important parts of circuit sizing: all circuit
specifications have to be satisfied for a solution (set of design parameters) to be feasible.
There are three
larger categories of constraint handling techniques \cite{li2018two}: techniques  that
place  a  higher  priority  on  feasible  solutions to  survive  to  the  next
generation,  techniques  for  balancing  the  trade-off between  feasibility   and
convergence,   and   techniques   for   repairing infeasible solutions.  We focus on the 
methods in the first category, as they are robust, well-understood and tested in literature. 

Firstly, the violation of a single constraint is usually evaluated as $max\{c_i(x),0\}$: how far
is the solution $x$ from satisfying constraint $c_i$. When the constraint is met, the
violation is $0$. We use the term \emph{Constraint Violation} (CV) for a solution as the
average  of the normalized violations of all constraints of that solution \cite{zeng2017general} :
\begin{equation}
    CV(x) = \dfrac{1}{N_c} \sum_{i=1}^{N_c} \dfrac{cv_i(x)}{\max\limits_{x \in P0}\{cv_i(x)\}}
    \label{eqn:CV}
\end{equation}
By normalizing based on the initial uniformly initialized population $P0$, the constraint
violation is fair for every constraint and does not require additional weight parameters
to be tuned. 

The idea that feasible solutions should be preferred to infeasible ones was introduced
and then used in NSGA-II \cite{DEB2000311,Deb2002-bp}. The following constrained dominance relation was
designed - a solution $s1$ dominates $s2$ if:
\begin{inparaenum}[1)]
    \item $s1$ is feasible but $s2$ is not,
    \item they are both unfeasible, but $s1$ has a smaller CV, 
    \item they are both feasible and $s1$ dominates $s2$.
\end{inparaenum}
This mechanism is widely used and led to a two-step strategy for solving constrained
optimization problems \cite{Venkatraman2005-jr}. The first step focuses only on
satisfying the problem constraints, by minimizing CV until it reaches $0$. The second
step then optimizes the objective functions.  

This method is robust, easy to implement and does not require parameter tuning, reason
for which it is arguably the most widely used. Assigning feasible solutions a higher
priority will speed up the convergence if the feasible region is not extremely difficult
to be found. However, prioritizing feasibility might results in poor objective values once
the feasible region is reached, and potentially reduce the population diversity
\cite{chen2019novel}. For these reasons, we choose to use this methodology to handle
constraints and implement extra mechanisms for maintaining the population diversity to
reduce the potential negative impact.

\subsection{Multi-Objective Performance Metrics}\label{subsec:HV}
The solutions provided by multi-objective optimization algorithms represents a set of
approximate Pareto optimal solutions. Evaluating the quality of such a set is usually
done on one or more of the following dimensions: 
\begin{itemize}
    \item \emph{Convergence}: how close the solutions are to the Pareto optimal solutions.
    \item \emph{Uniformity}: how uniformly distributed uniformly the solutions are along
    the Pareto optimal frontier.
    \item \emph{Distribution}: how well the solutions cover the whole Pareto optimal frontier.
\end{itemize}

Evaluation is often realized by assigning a score to a set of solutions, which summarizes
one of more of the above dimensions. The most common quality indicator is the Hypervolume
(HV) metric. 

\paragraph{Hypervolume (HV)} 
It requires a \emph{nadir} reference point, for example, the worst point in the objective
space found so far. HV measures the size of the objective space bounded by this nadir
point and dominated by the given solutions. The volume is computed using the Lebesgue
measure. 

The larger the HV value is, the better the quality of the solutions is at approximating
the PF. Moreover, HV is the only known indicator that reflects the Pareto
dominance in the sense that, if an approximation set weakly dominates other, this fact
will be reflected in the values of HV. It has also been shown that, given a finite the
search space, under certain conditions, maximizing HV corresponds to finding the optimal
Pareto set when the number of approximation points goes to infinity. In addition, it has
been experimentally observed that maximizing the HV leads to subsets of PF that present
high diversity levels. The drawback with this indicator is that its computational
complexity increases exponentially with the number of objective functions. For
many-objective optimization problems with more than 10 objectives Monte Carlo-based
sampling techniques are required to estimate it. 

Another quality indicator metric is the \emph{Hyperarea Ratio} (HR), which is 
defined as the ratio between the HV of the PF known so far and
the HV of the true PF. Nevertheless, this metric requires the researcher to know
the true PF. The \emph{Inverted Generational Distance} (IGD) represents another
popular metric which requires the true PF to be known beforehand. Our proposed
algorithm and the results discussions use HV as a quality indicator metric for the reasons outlined above.
 
\paragraph{Diversity}
It represents an important topic in multi-objective optimization, as we aim to provide results that
contain meaningful choices from which a designer can pick. Diversity measures the quality
of the approximate PF in terms of both uniformity and distribution. It also
plays an important part in the exploration-exploitation trade-off. Maintaining good
diversity during the optimization is necessary to ensure good exploration of the solution
space and to avoid getting stuck in local optima. 

Literature studies showed that (i) \emph{spacing} \cite{schott1995fault} and (ii)
\emph{Crowding Distance} (CD) \cite{Deb2002-bp} are good metrics for assessing the
diversity gain obtained by adding a new individual to a population \cite{Audet2018-ar}.
It is worth noting that these metrics are useful for creating a new population, step by
step, but cannot be used to derive a single number representing the diversity of a given
population.

On the other hand, \emph{Distribution Metric} (DM) is an effective metric for assessing
the diversity of a population \cite{Zheng2017-em}. DM evaluates both the spread and the
uniformity of the individuals and it is not computationally intensive (\emph{O(N\textsuperscript{2})}). 
DM computation is based on the distance between individuals projected individually on
each of the objectives. It takes into account the mean, the standard deviation and the
range of these projections and also the \emph{nadir} (worst possible) and the
\emph{ideal} (best possible) individuals. 

Compared to the methods referenced in this section, our algorithm uses this distribution
metric directly in one of the survival and one of the candidate selection methods
proposed. In these two methods the unfeasible solutions are not compared using only
CV, which might lead to loss of diversity. Instead, they are compared
on both CV and DM objectives, in an on the fly bi-objective optimization problem. This
promotes a natural balance between convergence and diversity, without the need of
additional hyper-parameters that need tuning. 

\subsection{Evolutionary Algorithms}\label{subsec:EAs}
Evolutionary Algorithms (EAs) are stochastic search methods that mimic the survival of the
fittest process of natural ecosystems. This type of algorithms, in relation to the
multi-objective optimization, are designed to meet two common but conflicting goals:
minimizing the distance between solutions and the PF (\ie convergence) and
maximizing the spread of solutions along the PF (\ie diversity). Naturally, balancing
convergence and diversity becomes much more difficult in multi-objective optimization.

There are many examples of successful EAs employed for circuit sizing. One of the most widely used
algorithm is NSGA-II \cite{Deb2002-bp}, used as-it-is for automatic analog IC sizing
in \cite{Lourenco2012-fm,Lourenco2016-na} or with the addition of a clustering method to
reduce the number of simulations to be performed  \cite{Canelas2017-qb}. Other popular
algorithms employed for circuit sizing are PSO and various Differential Evolution (DE)
variants \cite{Goudos2011-yd}. 

While EAs can produce design solutions of good quality, an important challenge is the
optimization efficiency with respect to runtime. Simulations can be extremely
computationally expensive and thus using default EAs can take an impractically long time
to produce the required results. For this reason, surrogate models are used to replace a
fraction of the expensive simulations. The construction of the surrogate model and its
use to predict the function values typically costs much less effort than embedding the
expensive function evaluator within the optimizer as-is.

\subsection{Surrogate Models}\label{subsec:surrmodels}

The most popular surrogate models used in the last several years are by far the GPs
 and artificial NNs, although other options have also been
used, such as response surface methods and Support Vector Machines
\cite{Voutchkov2010-iv}. 

Early work in this area is based mostly on \emph{offline training}: the model is fitted once on
many available datapoints and then used with very few updates during the optimization
process. For example \cite{6903368,6510530} combine GPs (only making use of the predicted
values, but not of the uncertainty information) with multi-objective search algorithms for
circuit sizing.

More recent methods, including our approach, use \emph{online training}: the model is
incrementally updated during the optimization process with each newly available
simulation. While the extra surrogate model operations add computational cost, they also
guide the selection of the future candidate points more efficiently towards faster
convergence. There are several methods that combine EAs with GPs
\cite{Liu2011-bu,liu2011da,Liu2012Bo,Liu2016, Liu2014} but they can only deal with single
objective (and multiple constraints) circuit problems. Moreover, \cite{Liu2012Bo} relies
on a stage-by-stage methodology with a complex implementation. All five methods prioritize feasible
over unfeasible solutions and use static penalty functions, for which in some instances
the weights are manually specified \cite{Liu2011-bu,liu2011da}. We believe that scaling
the CVs based on an uniform initial population is a more robust
approach. Another difference to our method is that these generate several offspring, but
only the child with the maximal acquisition value is selected as the next data point to be
explored. That does not allow for batches of simulations to be run in parallel efficiently, representing 
an important bottleneck when simulations are very expensive. 

The acquisition function used are \emph{Expected Improvement} (EI) \cite{Liu2011-bu,Liu2012Bo}, \emph{Lower Confidence Bound} (LCB)
\cite{Liu2014,Liu2016} or even a custom methodology using mean square error thresholds
\cite{liu2011da}. Although \cite{Liu2014,Liu2016} use LCB, the weight of the uncertainty
term is set to $0$ for the GP models of constraints, thus neglecting the uncertainty
information for constraints. 

\paragraph{Bayesian Optimization}
Other recent approaches fully rely on the BO approach
\cite{Lyu2018-mb,he2020}. They do not use an EA to generate candidates from which to
preselect based on a surrogate model, but choose the next point solely based on
optimizing the acquisition function. While this approach can be more efficient by
requiring fewer simulations overall, these will have to be executed sequentially. When
simulations are so expensive that only few tens of iterations can be afforded, we believe
a better result can be obtained by using EAs and evaluating several simulations in
parallel at each iteration. Moreover, we believe that the main advantage of novel
Bayesian methods such as sparse GPs 
\cite{he2020} is not the improved GP training and prediction speed, but mitigating
overfitting and we might consider a similar approach in future work. 
Furthermore, for circuits -- as in our case -- with design variables that are integer valued or
categorical, the assumption that the acquisition functions are differentiable over the
input space is no longer valid. Then, the acquisition functions cannot be
efficiently optimized via gradient descent anymore. EAs, on the other hand, are well
equipped to solve optimization problems without the differentiable requirement. 

In both \cite{Lyu2018-mb,he2020}, the EI is used as the acquisition function but the
value is also weighted by the probability of satisfying the constraints. One limitation
is that these two methods are based on the assumption that the constraints are independent so
that they can be independently modeled  and  the  probabilities could  be  multiplied,
which might not hold true. We prefer to use the LCB acquisition function and handle the
constraints via optimizing the CV. Furthermore, these two methods can
tackle multi-objective problems but do so by 
transforming them into single-objective ones via Chebyshev scalarization.
While better than a linear scalarization, this approach still has the drawback of defining
a bias a priori in terms of weights \cite{van2013scalarized}. In comparison, we search
directly into the multi-objective space using Pareto dominance rules and the HV
metric, without any scalarization. 

\paragraph{Multi-objective Acquisition Ensemble}
Another approach which also relies on BO is the Multi-objective Acquisition Ensemble (MACE) algorithm \cite{pmlr-v80-lyu18a}. While other BO
algorithms use a single acquisiton function, this method selects multiple acquisiton functions: LCB,
EI and \emph{Probability of Improvement} (PI). Then the algorithms performs
multi-objective optimization to find the PF of the acquisition functions and samples the PF to obtain 
multiple candidate points for the objective function evaluations. This way the best trade-off between acquisition functions
can be captured. For multi-objective problems such as circuit design optimisation, an objective function must be 
constructed from the circuit responses in order to use the BO algorithm on it.


\paragraph{Graph Convolutional Neural Networks - Reinforcement Learning}
A different approach in single-objective optimisation for Circuit Design is GCN-RL. The 
DRL based method first introduced in \cite{wang2018learning} and developed further in \cite{wang2020gcn} focuses 
on using $transfer$ $learning$ in conjunction with $Reinforcement$ $Learning$ and $Graph$ $Neural$ $Networks$. The GCN-RL agent 
extracts features of the topology graph whose vertices are transistors, edges are wires. 
Using those features, after training a DRL agent on a circuit, the same agent can be used 
on other topologies, where transfer learning significantly reduces the required number of simulations. 
Just as with MACE, GCN-RL can be used on single-objective functions in order to apply the algorithm
on circuit design, an objective function must be constructed from the circuit responses.



\section{The Proposed Method}\label{sec:method}

In this section we propose and present an innovative circuit sizing method further called 
Multi-objective Optimization based on Differential Evolution and Bayesian Inference (MODEBI).
The method is adapted to circuits with a high number of design variables (tens) and 
many PVT (process, voltage, temperature) corners.

\begin{figure*}[h]
    \centering
    \includegraphics[width=0.9\textwidth,clip]{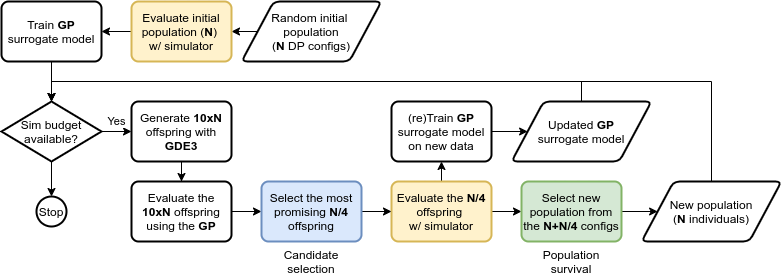}  
    \caption{Multi-objective Optimization based on Differential Evolution and Bayesian Inference}
    \label{fig:mehtod}
\end{figure*}

The proposed method builds on top of a differential evolution algorithm presented in subsection
\ref{sec:method:GDE3}, as shown in Figure \ref{fig:mehtod}.
It uses a GP as a surrogate model to minimize the number of simulations 
required for the optimization (subsection \ref{sec:method:GP}). 
The innovative candidate selection methods and population survival policies,
designed to improve convergence speed and population diversity 
are introduced in subsection \ref{sec:method:GDE3GP}.

\subsection{Generalized Differential Evolution 3}\label{sec:method:GDE3}

The core idea of our research was to use a multi-objective EA (MOEA) 
for finding the optimum in the hyperspace generated by the design variables and PVT corners.
Our previous work consisted in considering five state-of-the-art MOEAs and comparing them in order to decide which one is the
most suitable for the complexity level of the circuit sizing tasks:
Non-dominated Sorting Genetic Algorithm II (NSGA-II) \cite{Deb2002-bp}, 
Non-dominated Sorting Genetic Algorithm III (NSGA-III) \cite{Deb2014-xj},
General Differential Evolution 3 (GDE3) \cite{Kukkonen2005-mj}, 
Indicator Based Evolutionary Algorithm (IBEA) \cite{Zitzler2004-tg},
and Strength Pareto Evolutionary Algorithm 2 (SPEA2)  \cite{Zitzler2001-jm}. 
First, we performed hyper-parameter tuning for all
five algorithms on several synthetic benchmarks with comparable dimensionality to the target circuit
sizing tasks, followed by the usage of the best configuration found for each one of them on the circuit sizing task. 
The extensive results can be found in \cite{Stn2021CAS} and \cite{Vis2021SpeD}. 

The obtained results highlighted that SPEA2 and NSGA-II did not perform as well as the other three methods. 
Furthermore, we found that compared to the other algorithms, GDE3 preserves a more diverse
population at the cost of
a slightly slower convergence rate. Specifically, GDE3 is showing a good exploratory behavior,
being less prone to getting stuck in a local optima. Moreover, GDE3 tends to find solutions of
similar quality to the ones found by the other  algorithms even in the early stages of
the optimization. The only drawback is that some lower quality solutions might be preserved
in the population for multiple epochs, resulting in a lower average
quality of the population in the early stages. However, given the high complexity of the
circuit sizing problems, a slower but more robust convergence is desirable. For these
reasons, we chose GDE3 as the base of our algorithm.

GDE3 modifies the selection rule of the basic \emph{Differential Evolution} (DE) and it extends
it for constrained MOOPs. The recombination corresponds to the most common DE operator,
DE/rand/1/bin \cite{Kukkonen2005-mj}. It 
involves creating an offspring solution $\vec{O}_{g,i}$ (\ref{eqn:offspring}) by
replacing one or more input variables of an existing "parent" $\vec{P}_{g,r1,i}$ in the
population of the current generation $g$ (\ref{eqn:recombination}). The replacement is
obtained by adding the weighted difference between two randomly selected solutions
$\vec{P}_{g,r2,i}$ and $\vec{P}_{g,r3,i}$ (\ref{eqn:replacement}).

\begin{equation}
    \vec{O}_{g,i} = (o_{g,i,1}, o_{g,i,2}, ..., o_{g,i,D})
    \label{eqn:offspring}
\end{equation}
\begin{equation}
    o_{g,i,j} = p_{g,r1,j} + F \cdot (p_{g,r2,j} - p_{g,r3,j})
    \label{eqn:replacement}
\end{equation}
\begin{equation}
    o_{g,i,j} = 
    \begin{cases}
        o_{g,i,j}, & \textrm{if } rand < CR \textrm{ or } j = j_{rand}\\
        p_{g,i,j}, & \textrm{otherwise}
    \end{cases}
    \label{eqn:recombination}
\end{equation}

The \emph{Crossover Rate} ($CR$) represents the probability of replacement, while
$j_{rand}$ ensures that at least one input variable is replaced. The $CR$ and
step size $F$ are the control parameters of GDE3.

\begin{algorithm}[h!]
    \KwIn{A random population \textbf{POP} of size $\textbf{N}$}
    \While{simulation budget available}{
        offspring = generate(POP, GDE3Operator)

        simulate(offspring)

        \For{$i=0$; $i<N$; $i++$}{
            POP[$i$] = paretoCompare(POP[$i$], offspring[$i$])\;}

        POP = nondominatedPrune(POP,$N$)
    }
    \caption{GDE3 for circuit sizing}
    \label{algo:GDE3}
\end{algorithm}

If the offspring solution performs better than the parent solution, the later will be
replaced by the former in the next population. In MOOPs there is a chance the offspring
and the parent do not dominate each other. In this case, both solutions are kept to avoid
expensive comparisons. Consequently, the population size can increase in time. On such occasions,
 the population is pruned back to its original and constant size using Pareto
non-domination. Solutions of the same Pareto rank are discriminated via CD. The pseudocode for GDE3 is shown as Algorithm \ref{algo:GDE3}.




\subsection{Gaussian Process Model}\label{sec:method:GP}

The GP regression is based on Bayesian inference, in which a prior
statistical model can be combined with observed evidence to deduce a more accurate
statistical model. A prior distribution can be specified using a prior mean function and
a kernel function. In the absence of prior knowledge, the mean is assumed to be zero. The
kernel function represents a measure of the similarity between the function values at two
locations. Training the GP over the available data points means encoding the information
in the mean and the kernel. 

For each new point, the GP will predict the mean ($\mu$), representing the estimate of
the function value and the variance ($\sigma^{2}$), representing the uncertainty of the
prediction. These predictions are refined incrementally when new data is observed. 


An acquisition function is typically used to explore the state space based on the GP
model. Various types of acquisition functions can be used to balance the exploration and
exploitation during the optimization. In the proposed algorithm, we employ the lower confidence bound: 
\begin{equation}
    LCB(x) = \mu(x) - K\sigma(x) 
\end{equation}

In minimization problems, a larger $K$ corresponds to a highly exploratory behavior.

As the predictions are based on the correlation among the observed evidence and not on
 the intrinsic structure of the data, GP models are less prone to overfitting compared to
 other methods \cite{Liu2012-cr}. Furthermore, we observed empirically that GP models
 perform significantly better than NN-based models, when a smaller number of
 data points are available (fewer than $5,000$, as our problem of circuit sizing requires). This results in higher
 accuracy, especially for the first steps of optimization when little data is available. 
 The drawback of GP models is the high computational cost which comes from the complexity
 of training and predicting, \emph{O(N\textsuperscript{3})} and \emph{O(N\textsuperscript{2})}
 respectively \cite{he2020}. However, this disadvantage can be partially overcome via
 parallelization, training data point selection or other techniques such as sparse GPs 
 \cite{he2020}.

In circuit sizing tasks, the optimization hyperspace tends to be complex due to the high
number of design parameters and PVT corners. In online approaches it
is crucial to have a reliable surrogate model from the start of the optimization,
when the training data set is quite small. Thus, the GP models are probably the most
common choice of online surrogate model in circuit sizing tasks
\cite{Liu2012-cr,Liu2014-kd,Lyu2018-rc,Huang2019-ib} and will be
used in our algorithm as well to speed up the optimization procedure.


\subsection{Multi-objective Optimization based on Differential Evolution and Bayesian Inference (MODEBI)}
\label{sec:method:GDE3GP}

\paragraph{Population Diversity} It represents a critical aspect in MOEAs not only because of the risk of
ignoring certain areas of the input space, but also because of the final goal of finding
multiple points on the Pareto front. The original GDE3 algorithm has a good mechanism of
preserving the diversity of the population, by comparing each offspring only with its
parent in order to decide which solution will be part of the next generation. However, integrating
the use of the GP surrogate model introduced a few changes to the original GDE3 approach
and extra incentive for diversity preservation was required. 
%
The original GDE3 algorithm uses CD in the pruning phase. For the
reasons presented in section \ref{subsec:HV}, we use instead the DM to monitor the diversity across the optimization, 
in addition to its involvement in
the selection of suitable individuals. More details will be provided below in the
description of the selection and survival methods. 

Having an intrinsic mechanism of preserving diversity, GDE3 is a good choice to sample a
high dimensional space to solve complex problems. Yet, its slow convergence can
represent an important drawback in problems where the evaluation is expensive, such as
circuit sizing. To overcome this issue, a GP is used as a surrogate model to minimize the
number of simulations required for the optimization.

\paragraph{Preselection}
We employ the GP model as a preselection mechanism. To maximize the potential benefit, 
the GDE3 offspring generator is used ten times, to produce ten times more offspring ($10N$) than
the population size ($N$).
At the same time, we want to reduce the number of simulations performed at each epoch, to
a quarter of the population size ($N/4$ instead of $N$ as in Algorithm \ref{algo:GDE3}).
Thus, we use the surrogate model to select the most promising $N/4$ offspring out of the
$10N$ generated. By replacing at most a quarter of the population each step, 
MODEBI finds a good balance between exploration and exploitation. The exploratory behavior of the
evolution is encouraged by maintaining a high number of individuals in the population. At
the same time, only the offspring showing important improvement rates are considered to
join the next population.

One GP is trained for each circuit response, with  the design parameters and the
operating conditions (OC) as inputs. The advantage of this approach is the lower
computational cost and the possibility of training and querying the GPs in parallel. One
drawback is that the correlations between responses are not taken into consideration,
which we plan to address in future work. We use a scaled \emph{Radial-Basis Function} (RBF) kernel.
Apart from being a popular choice for machine learning, its reduced complexity makes 
model training feasible on all available data points (up to $20000$ in our testcases).
Good accuracy is achieved with a reasonable computational cost.

The most promising $N/4$ offspring (as predicted by the GP and selected using one of the selection
functions introduced below) are evaluated using the
circuit simulator. Finally, the new population is created by choosing the best
individuals from the last population and these offspring, using one of the survival
functions described below. The updated method is presented
as Algorithm \ref{algo:GDE3GP}.

\begin{algorithm}[h!]
    \KwIn{A random population \textbf{POP} of size $\textbf{N}$}
    \While{simulation budget available}{
        GP = train(available simulations)

        offspring = $10 \times $generate(POP, GDE3Operator)

        evaluate(offspring, GP)

        bestOffs = select(offspring, $N/4$)

        simulate(bestOffs)

        POP = survival(POP, bestOffs, $N$)
    }
    \caption{MODEBI for circuit sizing}
    \label{algo:GDE3GP}
\end{algorithm}

Generating $10N$ offspring and replacing only $N/4$ individuals in the population has its benefits,
as explained above, but comes with additional complexity.
The original GDE3 offspring selection and survival procedures cannot be used unaltered.
We propose two strategies for each of these procedures, as described below.

\paragraph{Offspring Selection}
While in vanilla GDE3 the unique offspring of a parent was compared only to its parent and 
the better individual was promoted,
in our proposal we have $10N$ offspring out of which we would like to select $N/4$.

\textit{Hereditary Selection} (\beastEach), our first proposal,
is inspired from the original GDE3 selection mechanism and is designed to preserve the diversity of the population.
\beastEach \hspace{0.5pt} works in three steps (see Algorithm \ref{algo:HSel}):
(i) compares the offspring of each parent and selects the best one,
(ii) scores the $N$ offspring against their parents and
(iii) selects the $N/4$ offspring that score best at step (ii).

At step (i) the offspring of each parent are compared in terms of performance, and one winning
offspring is selected per parent:
\begin{itemize}
    \item a feasible offspring is preferred to an unfeasible one;
    \item amongst feasible offspring, the one which would add the most HV to the
    current population is chosen;
    \item if all offspring are unfeasible, the one with the lowest CV
    (subsection \ref{subsec:metrics}) is chosen.
\end{itemize}

Step (ii) assigns a score to each of the $N$ winning offspring based on the relative
improvement -- if any -- to its parent:
\begin{itemize}
    \item if both the offspring and the parent are feasible, the score is the difference
    between the HV contribution of the offspring and the parent to the
    population (minus the parent in question); 
    \item if both the offspring and the parent are unfeasible, the score is $1 +$ the
    difference between the parent and the offspring's CV; 
    \item if the parent is unfeasible and the offspring is feasible, the score is $10 +$
    the above CV difference. 
\end{itemize} 

Finally, at step (iii) the $N/4$ offspring with the best improvement scores are considered to be the
most promising ones and selected for simulator evaluations. This scoring method will
promote solutions meeting the constraints first, then optimizing the objectives. 

\beastEach \hspace{0.5pt} preserves the diversity of the population 
by allowing only one offspring per parent to be promoted in the next population.
However, it might be the case that several offspring generated by the same parent are very good and also diverse.
To effectively address this case and allow more offspring per parent to be promoted, 
we also introduce \textit{Pooled Selection} (\bestPooled).

\begin{algorithm}[h!]
    \KwIn{Current population \textbf{POP} of size $\textbf{N}$ and its offspring of size $\textbf{10N}$}
    \KwOut{bestOffs of size $\textbf{N/4}$}
    \For{$i=0$; $i<N$; $i++$}{
        bestOffs[$i$] = compareOffspring(offspring[parent=POP[$i$]])
    }
    \For{$i=0$; $i<N$; $i++$}{
        score[$i$] = computeImprovement(POP[$i$], bestOffs[$i$])
    }
    bestOffs = select(bestOffs[max score], $N/4$)
    \caption{Hereditary Selection Algorithm}
    \label{algo:HSel}
\end{algorithm}

\bestPooled \hspace{0.5pt} is directly
selecting the best $N/4$ offspring out of the $10N$ without using any parent information. 
As illustrated in Algorithm \ref{algo:PSel}, 
\bestPooled \hspace{0.5pt} selects $N/4$ feasible offspring one by one based on their HV,
if the total number of feasible solutions is bigger than $N/4$. The HV computation takes into 
account the solutions selected at previous steps and also a fraction of the best solutions in the 
current population. Thus, at every step the solution adding more HV is selected. The 
number of considered solutions from the current population can be selected by the user. This 
parameter balances the selection duration and its accuracy. The best case scenario is when all
of the current population is used, but for problems with a high number of objectives this would 
be too computationally expensive.
If the number of feasible solutions is smaller than $N/4$, it selects all feasible offspring 
and the best unfeasible offspring in terms of CV and DM, with regard to the current population.
A similar strategy outlined below as \textit{Improved Survival} is also used for population survival.

\begin{algorithm}[htb!]
    \KwIn{Current population \textbf{POP} of size $\textbf{N}$ and its offspring of size $\textbf{10N}$}
    \KwOut{bestOffs of size $\textbf{N/4}$}
    bestOffs = None

    \For{$i=0$; $i<10N$; $i++$}{
        CVs[$i$] = computeConstraintViolation(offspring[$i$])}
    feasOffs = select(offspring[$CVs=0$])

    \If{feasOffs $> N/4$}{
        bestOffs = []

        consPOP = select(POP[max HV], max\_pop)

        \While{size(bestOffs) $< N/4$}{
	        feasOffsHvs = computeHV(consPOP + bestOffs, feasOffs)

            bestOffs += select(feasOffs[max feasOffsHvs], $1$)
        }
    }
    \Else{
        bestOffs = feasOffs

        unfeasOffs = select(offspring[$CVs>0$])

        \While{size(bestOffs) $< N/4$}{
            DMs = computeDM(POP, bestOffs, unfeasOffs)

            bestOffs += select(unfeasOffs[min CV and DM], $1$)
        }
    }
    \caption{Pooled Selection Algorithm}
\label{algo:PSel}
\end{algorithm}

\paragraph{Population Survival}
After the best $N/4$ offspring are selected as described above and 
simulated (line 6 in Algorithm \ref{algo:GDE3GP}),
a population survival strategy must be applied to create a new population (size $N$) from the $N+N/4$ individuals.
We further propose two such strategies.

\textit{Pareto Survival} (\ParetoSurvival) is based on the original GDE3
selection mechanism and can only be applied if \beastEach \hspace{1pt} selection was performed.
The new offspring and its parent are directly compared in terms of Pareto dominance, and the
solution that dominates the other replaces the dominated one in the new population. If
the two are on the same Pareto rank, they are both kept. Moreover, the $3N/4$ parents
without selected offspring will also be added to the new population. To maintain the
population size constant, the new population is pruned back to $N$ members using  Pareto
non-domination, similarly to vanilla GDE3. 

\textit{Improved Survival} (\newSurvival) is more complex and uses the HV
and DM metric to select the best solutions for the new population:
\begin{itemize}
    \item feasible solutions are preferred to unfeasible ones;
    \item if there are more feasible solutions than $N$, a fraction of solutions
    producing the highest HV will be selected. Then, the rest will be chosen
    using Crowding Distance. Ideally, all the solutions would be chosen based on HV,
    but that is computationally unfeasible when the number of objectives is high.
    \item if only unfeasible solutions are left and some of them have to be chosen.
    They will be selected iteratively by always picking the best solution in terms 
    of CV and DM, with regards to the solutions already selected. 
\end{itemize}

\paragraph{Summary}
MODEBI can be regarded  as an extension of GDE3 incorporating a GP surrogate model to preselect 
candidates in order to speed-up convergence without wasting time on real circuit simulations.
The preselection needs to be done carefully in order to preserve the population diversity and 
promote the most promising offspring.
To achieve this the method includes new candidate selection (\beastEach, \bestPooled) and 
survival (\ParetoSurvival, \newSurvival) strategies that are expected to 
promote population diversity (\ParetoSurvival-\beastEach),
offspring performance (\newSurvival-\beastEach) or
even both (\newSurvival-\bestPooled)



\section{Experimental Results}\label{sec:results}

In this section, we present the two circuits used for the experiments (subsection \ref{sec:results:circuit}), as well as 
the results obtained with baseline Evolutionary Algorithms (subsection \ref{sec:results:baselineGDE3}),
the results obtained with baseline Bayesian Optimization algorithms (subsection \ref{sec:results:BO}),
and the results obtained when employing various scenarios of the proposed method - MODEBI (subsection \ref{sec:results:MODEBI}).
Finally, in subsection \ref{sec:results:comparison} we compare directly the most promising EA - GDE3, 
the best BO algorithm - MACE, and the proposed method - MODEBI, while in subsection \ref{sec:results:randomness} 
we evaluate the randomness impact on the various scenarios of MODEBI.

\subsection{Circuits Description}\label{sec:results:circuit}

Two circuits were used to run the experiments proposed in this paper. The first circuit is a 
voltage regulator with 27 design variables, including transistor
sizes, resistances, capacitances and the bias current.
There are 17 real design variables and 10 integer design variables. 
The circuit needs to meet the specifications in eight PVT corners.

The objectives to be optimized for the voltage regulator are Power Supply Rejection Ratios (PSRRs)
measured at various frequencies: 100k, 1M and 2M. 
Besides these three objectives, we monitor eight other responses, such as 
the phase margin, the deviation of the output voltage, integrated noise at various frequencies,
output current, for which the circuit needs to meet specific constraints.
In order to be considered feasible, a solution (i.e. a tuple of design variables) should meet all the constraints
in all PVT corners.

After finding feasible solutions, the goal of the circuit sizing method is to maximize the three objectives 
mentioned above. The budget is limited to $15,000$ simulations due to time constraints for the overall circuit 
sizing problem (e.g. each iteration of circuit sizing for a fixed topology should take at most 1 day).

As for the second circuit considered in this paper, it represents a 2.5V regulator having eight design parameters 
and six circuit responses with their associated constraints; all of them must be optimized in ten PVT corners. 
The design variables include channel widths, multiplication factors and capacitances, 
out of which six are real and two are integer design variables.

Similarly, the objectives to be optimized (more specifically maximized) for the second circuit are also PSRRs measured at 
different frequency intervals, as follows: [1k,10k], [10k,100k], [100k,500k] and [500k,3M]. 
The maximum PSRR for each interval is returned during each circuit simulation.
Apart from these, the quiescent current and the power stage and capacitors area have been considered 
and must be minimized by the algorithm. 

The feasible solution definition remains consistent - 
all constraints must be respected in all PVT corners. Next, the algorithm continues the optimization 
(maximization or minimization, depending on the objective constraint) by trying to find solutions with 
the best possible responses, without any of them falling out of the specifications. The simulation budget 
granted by the circuit designer involved $20,000$ circuit simulations, equivalent to a 2-day overall 
sizing iteration of the circuit. 

It must be emphasized that the proprietary simulator used for the two circuits allows running simulations in parallel. 
Therefore, the simulation budget for both circuits were spent in batches of $50$ parallel simulations. 

In the following subsection, the evaluation and comparison of the various algorithms 
is performed using the following performance figures:
\begin{inparaenum}[1)]
    \item the CV of a solution, described in subsection \ref{subsec:metrics} and
    \item the HV of a population of solutions, described in subsection \ref{subsec:HV}. 
\end{inparaenum}
While HV can be used to compare populations of feasible solutions, 
we employ CV to measure how far a solution is from meeting the constraints of the problem.


\subsection{Evaluation of baseline Evolutionary Algorithms}\label{sec:results:baselineGDE3}
As highlighted in section \ref{sec:method:GDE3}, our previous work published in \cite{Stn2021CAS} and \cite{Vis2021SpeD}
involved a baseline evaluation of the most promising five EAs (NSGA-II, NSGA-III, GDE3, IBEA, SPEA2) in the context of 
circuit sizing in a multi-objective optimization scenario 
for the two circuits presented in the previous section.

\begin{figure*}[h]
    \centering
    \includegraphics[width=1\textwidth,clip]{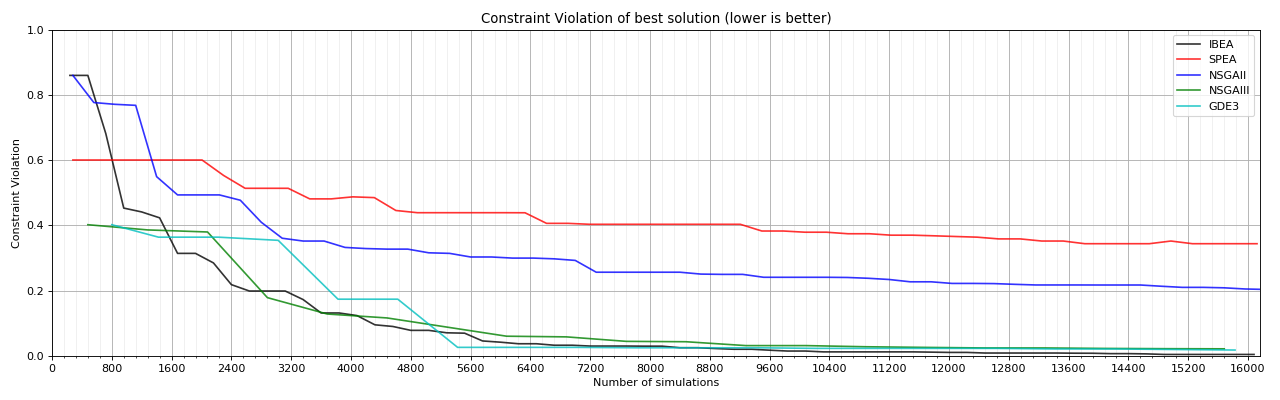}  
    \caption{The baseline evaluation in terms of CV of the five evolutionary algorithms on circuit 1. Image from \cite{Stn2021CAS}.}
    \label{fig:CAS_5Algs}
\end{figure*}

Figure \ref{fig:CAS_5Algs} depicts the CV of the best solution in each population obtained when applying the 
algorithms on the first circuit, based on $16,000$ simulations, as previously published in \cite{Stn2021CAS}. 
It must be recalled that the CV of a solution represents the average of the normalized violations of all constraints 
of that solution and it reaches zero when the solution becomes feasible. Although no algorithm was able to find a 
feasible solution, it can be easily observed that better solutions are reached as the population evolves. 

NSGA-III, IBEA and GDE3 reach substantially better results. Nevertheless, IBEA introduces significant computational 
costs for problems with a large number of objectives. NSGA-III and GDE3 display similar behavior, but GDE3 presents the 
advantage of working better with a larger population size, leading to better solution diversity. 
At the same time, the chance of getting stuck in a local optimum is highly reduced.
Although GDE3 has the lowest evolution capacity, as the number of evolutionary iterations of an algorithm is 
inversely proportional to the population size, this disadvantage is overcome by the featured advantages.

\begin{figure*}[h]
    \centering
    \includegraphics[width=1\textwidth,clip]{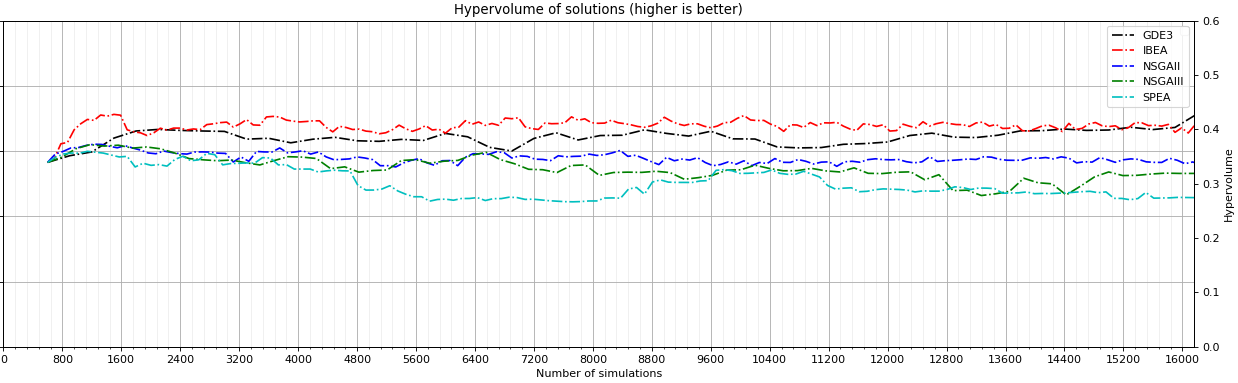}  
    \caption{The baseline evaluation in terms of HV of the five evolutionary algorithms on circuit 2. Image from \cite{Vis2021SpeD}.}
    \label{fig:SpeD_5Algs}
\end{figure*}

Figure \ref{fig:SpeD_5Algs} illustrates the HV computed on a set that contains the best 30 solutions returned by each algorithm,
as presented in \cite{Vis2021SpeD}. In this way, there is no longer an intrinsic advantage present for the algorithms with
 higher populations. GDE3 and IBEA proved to be better than the other three, in what concerns both solution quality and diversity.

Correlating these results, we can conclude that GDE3's performance is the best in all the considered scenarios and the algorithm 
represents the most promising among the state-of-the-art EAs for circuit sizing. The algorithm displayed versatility for circuits with 
relatively large number of objectives to be optimized, as well as a high capacity in what regards solution diversity, 
even when coping with a large population. However, supplementary effort is required to improve its slower convergence 
introduced by larger population sizes.

\subsection{Evaluation of the Bayesian Optimization Algorithms}\label{sec:results:BO}
It has been previously highlighted that EAs introduce a high computational expense in terms of required simulations.
In order to overcome this, the next step of our analysis concentrated on evaluating two of the most auspicious
surrogate models-based algorithms: BO and MACE (see subsection \ref{subsec:surrmodels}). 

Since the ultimate aim of our analysis is to establish a comparison benchmark for the introduced method (MODEBI), 
it is mandatory to be able to compare the best BO-based algorithm with the most promising EA, namely GDE3. Therefore, 
we established an optimization methodology, taking into consideration that BO algorithms are single-objective. For both 
algorithms - BO and MACE - we employed 2 GPs for training, in a sequential manner. The first GP is used 
and trained at each iteration, in order to minimize CV as an objective function. Thus, it is trained until a feasible 
solution is reached. While the first GP is used, data for training the second GP is saved.
Then, after finding a feasible solution, the second GP is trained using current and previously saved data at each 
iteration. The employed objective function for the second GP is expressed in the following equation; this way, 
we encourage the algorithms to maximize the HV of the solution while keeping CV close to 0:

\begin{equation}
    f_{objective} = 5CV - HV.
\end{equation}

By using this methodology, we can compare both surrogate model-based algorithms, BO and MACE, to the GDE3 and next 
to our proposed algorithm, in terms of the same criteria: number of simulations, time to find the first feasible  
solution and quality of the solutions, as well as the diversity of solutions after finding feasible ones. Unlike the EAs, 
BO and MACE do not require an initial population of hundreds of solutions. Both algorithms use a random
starting population of only five solutions for the initial GP training.

\begin{figure*}[h]
    \centering
    \includegraphics[width=1\textwidth,clip]{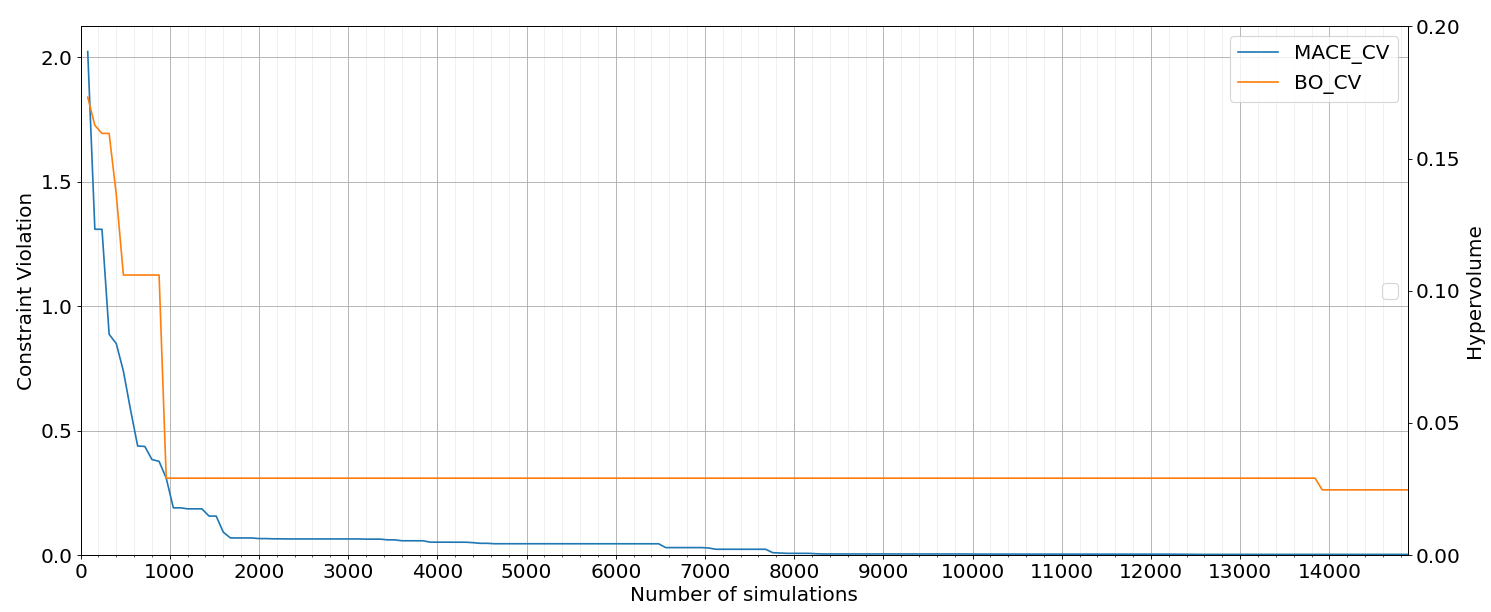}  
    \caption{The evolution of the CV of BO and MACE for the best solution among the $15,000$ simulations 
    budget on circuit 1}
    \label{fig:MACEvsBOB1}
\end{figure*}

For that reason, BO and MACE were evaluated in the context of circuit sizing in a multi-objective optimization 
scenario for the two circuits presented in the previous section, based on the presented methodology. Figure 
\ref{fig:MACEvsBOB1} illustrates the CV of the best solution found so far while using the $15,000$ simulations budget.
In this experiment the two algorithms were used to optimize the first circuit. 

We can easily observe that MACE outperforms BO; although MACE shows a slow convergence for the first $1,000$ simulations, 
it is able to reach a solution with CV$\sim0.0$ after roughly $7,000$ simulations. Whereas BO acquires the best solution 
with a CV$\sim0.25$ based on more than $14,000$ simulations (double). Unfortunately, even with its faster convergence speed, 
MACE was not able to find feasible solutions for this difficult problem. 

\begin{figure*}[h]
    \centering
    \includegraphics[width=1\textwidth,clip]{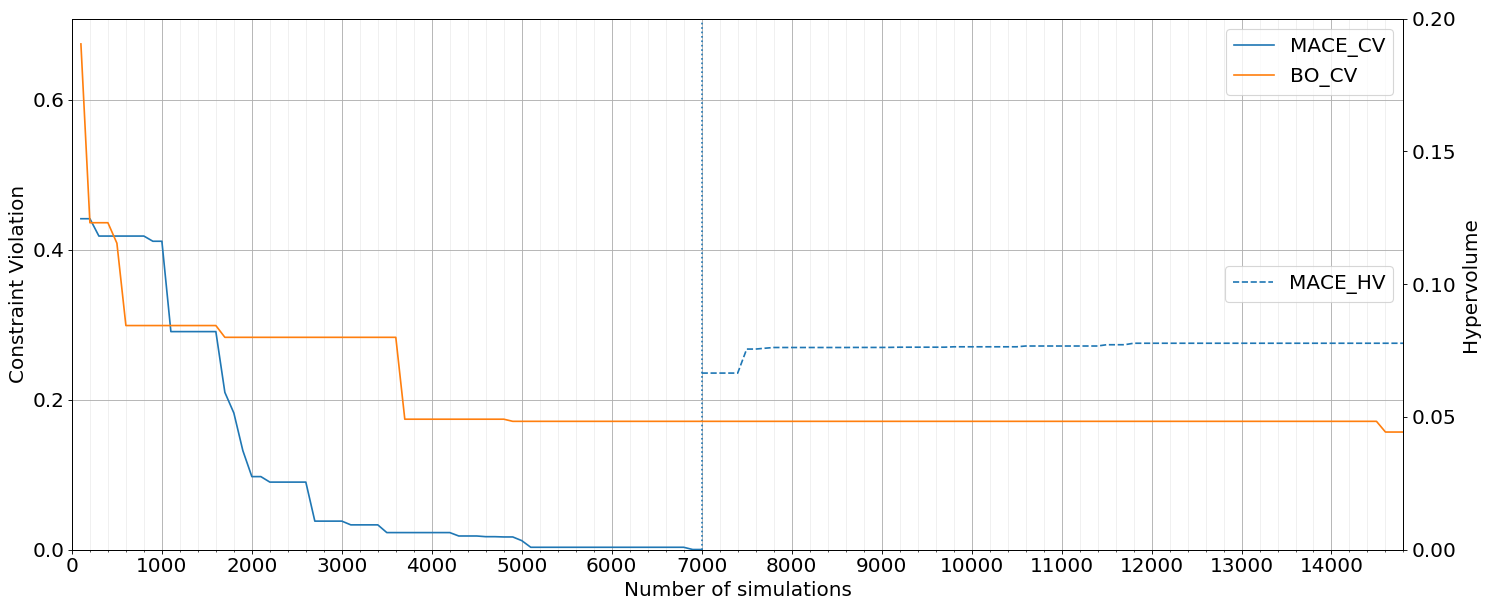}  
    \caption{The evolution of the CV and HV of BO and MACE for the best solution among the $15,000$ simulations 
    budget on circuit 2}
    \label{fig:MACEvsBOL1}
\end{figure*}

When applying the same methodology of the second circuit, we obtain the results presented in Figure \ref{fig:MACEvsBOL1}. 
First, MACE manages to find a feasible solution for this quite complex circuit, in contrast to BO and its 
slightly faster convergence (the orange line). MACE reaches CV$=0.0$ after less than $7,000$ simulations, comprising
only feasible solutions (the solid blue line), therefore the vertical dashed line marks the first feasible solution found by MACE.

Consequently, more than half of the available simulation budget is spent for improving the HV of the solution 
(depicted as the dashed horizontal line). Finally, the best HV is set 
around $0.075$, revealing a good quality of the six objectives optimized for the employed circuit. 

Considering all the above, we can conclude that MACE represents a suitable surrogate model-based algorithm
for circuit design optimization. By identifying the best trade-off between the available acquisition functions, 
the algorithm reached feasible solutions for one of the two circuits under test. 

\subsection{The Proposed Method (MODEBI) Results}\label{sec:results:MODEBI}
In order to assess the MODEBI algorithm performance in what regards the circuit optimization of the two 
available circuits, we performed three independent experiments, under the form of three application 
scenarios of the proposed algorithm (as presented in subsection \ref{sec:method:GDE3GP}). Since using a GP model 
as a surrogate for the circuit simulator should allow us to obtain faster the feasible solutions, the aim of the scenarios 
was deciding which combination of offspring selection methods and population survival approaches leads to better 
results (in terms of HV and CV). 

The first scenario uses MODEBI with \ParetoSurvival \hspace{0.5pt} survival policy and 
the proposed \beastEach \hspace{0.5pt} offspring selection method (MODEBI-S1). 
The second scenario uses the \bestPooled \hspace{0.5pt} offspring selection method in order to promote 
the best offspring, while slightly neglecting the diversity of the population by employing 
\newSurvival \hspace{0.5pt} approach for population survival (MODEBI-S2). The third 
scenario preserves the same offspring selection method (\beastEach) as MODEBI-S1, but uses 
\newSurvival \hspace{0.5pt} survival policy (MODEBI-S3).

\begin{figure*}[h]
    \centering
    \includegraphics[width=1\textwidth,clip]{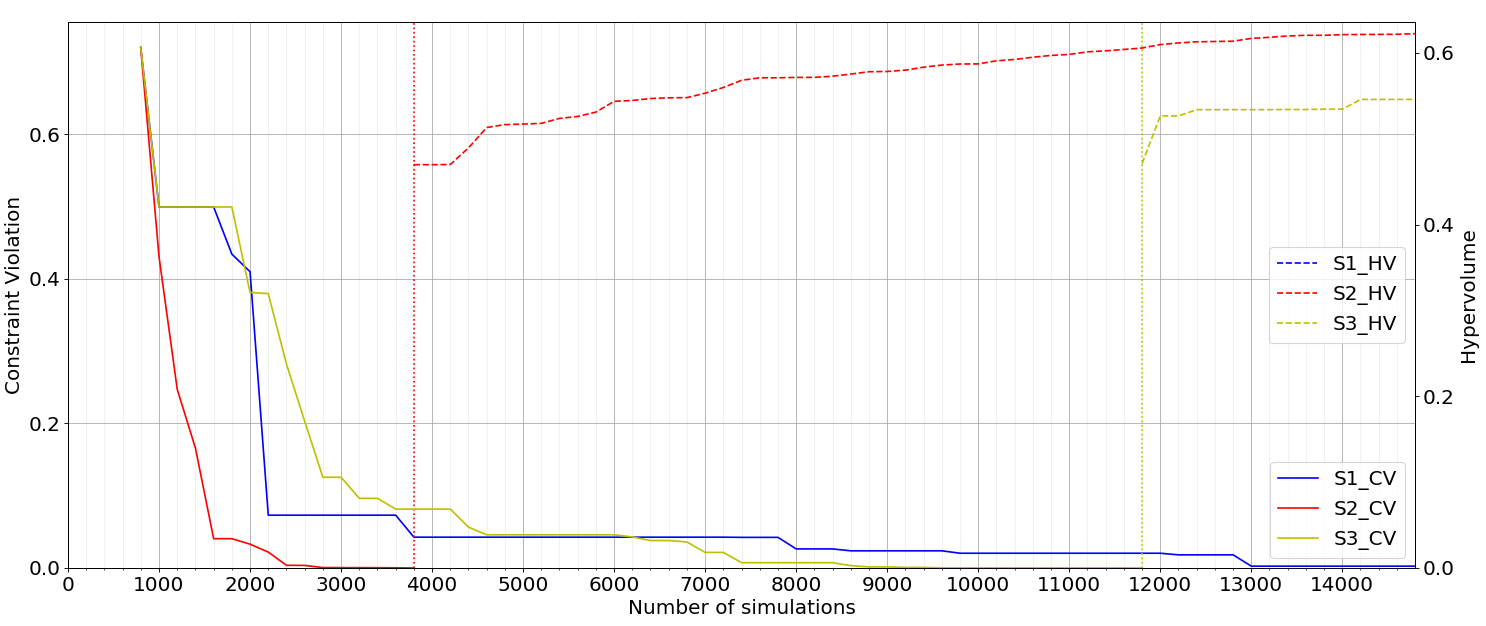}  
    \caption{The evolution of the CV for the best solution in the population (solid line) 
	and the evolution of the population HV (dashed line) for populations with feasible solutions for the
    three MODEBI scenarios on circuit 1}
    \label{fig:MODEBIB1}
\end{figure*}

Figure \ref{fig:MODEBIB1} summarizes the experimental results when applying the three proposed versions of 
MODEBI for circuit sizing on the first available circuit. During the first scenario, 
the proposed algorithm MODEBI-S1 (the solid blue line) does not manage to find 
good solutions at the beginning, although it converges fast. For example, a solution with CV$\sim0.05$ is 
obtained after less than $4,000$ simulations. It must be highlighted that the slow convergence for the 
first $2,000$ simulations is caused by the small amount of data available to train the GP.
However, after the GP is trained on more than $2,000$ datapoints its usage is evidently beneficial.
Unfortunately, even with its faster convergence speed, MODEBI-S1 was not able to find 
feasible solutions for this difficult problem.

In the second scenario, the combination \newSurvival-\bestPooled \hspace{0.5pt} leads to a huge boost in convergence speed (red solid line). 
MODEBI-S2 obtains a feasible solution after only $3,800$ simulations.
This way, more than $10,000$ simulations (dashed red line) are available for improving the HV of the population
(\ie working on both maximizing the objectives and obtaining diverse solutions). From this point further, 
we start monitoring the HV of the feasible members of the population in order to assess the quality 
of the three objectives. 

In scenario 3, MODEBI-S3 (solid yellow line) is able to find feasible solutions using the $15,000$ 
simulations budget, by employing the improved survival policy (\newSurvival). In this scenario, the proposed algorithm 
displays a faster convergence speed when compared to MODEBI-S1 and achieves a first 
feasible solution after less than $12,000$ simulations. Just as the case of MODEBI-S2, after obtaining the first 
feasible solution, the HV of the population for MODEBI-S3 is further monitored (dashed yellow line). 

In terms of HV, the comparison is performed between scenarios 2 and 3 (red and yellow dashed lines), where 
feasible solutions have been found. The second version of the algorithm, MODEBI-S2, obtains the best 
HV at the end of the optimization - more than $0.6$. One possible explanation of this outcome it that MODEBI-S2 
has at its disposal a much higher number of simulations remaining from the available budget for 
the optimization, unlike the third scenario, which achieves a HV of only $0.55$ using the remaining $3,000$ simulations.

\begin{figure*}[h]
    \centering
    \includegraphics[width=1\textwidth,clip]{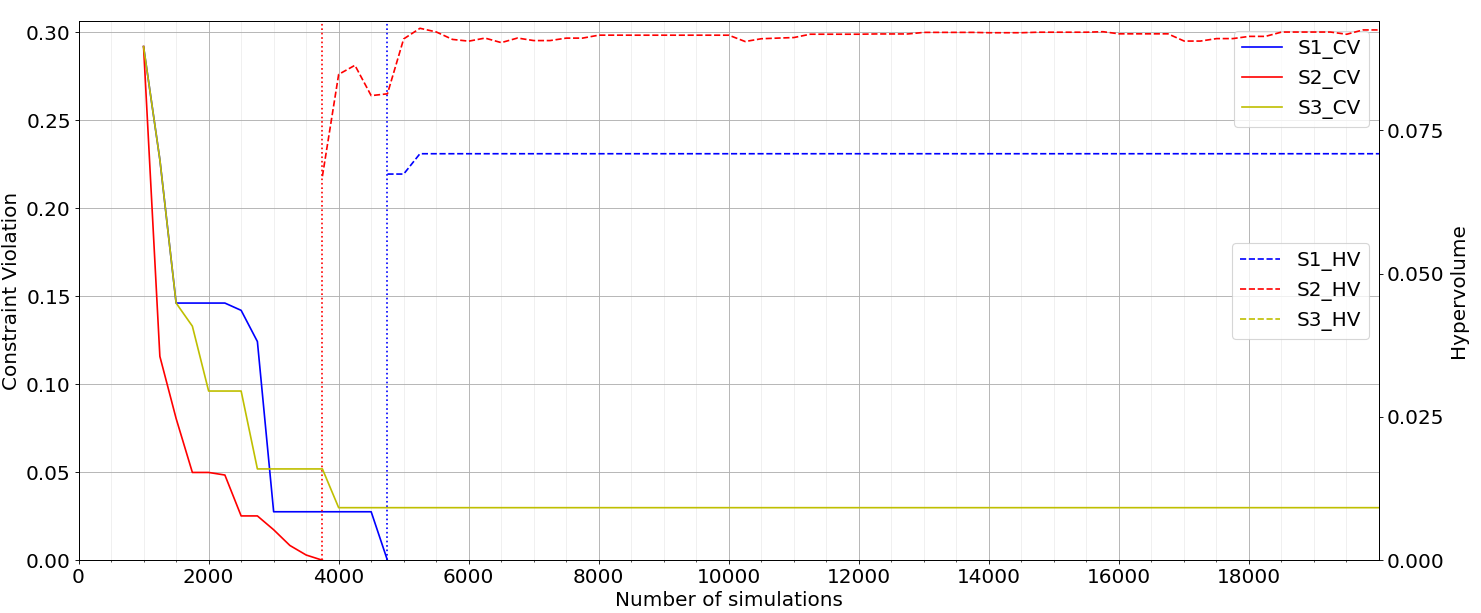}  
    \caption{The evolution of the CV for the best solution in the population (solid line) 
	and the evolution of the population HV (dashed line) for populations with feasible solutions for the
    three MODEBI scenarios on circuit 2}
    \label{fig:MODEBIL1}
\end{figure*}

The experimental results obtained on the second circuit are depicted in Figure \ref{fig:MODEBIL1}. For this circuit, 
MODEBI-S3 (solid yellow line), although characterized by a faster convergence (a CV$\sim0.05$ is reached after roughly 
$2,500$ simulations), was not capable to obtain feasible solutions. The best solution identified by this version has 
CV$=0.03$ after the simulation budget completion.

Contrary to circuit 1, the combination \ParetoSurvival-\beastEach\hspace{0.5pt} displayed by MODEBI-S1 (solid blue line) proved to be 
successful on circuit 2. Even though its convergence speed is slower than MODEBI-S3, it achieves the first feasible 
solutions after less than $5,000$ simulations. Afterwards, the rest of $15,000$ simulations are used to improve the HV of 
the population (dashed blue line), leading to HV$\sim0.07$. 

Once again, in the second scenario, the combination \newSurvival-\bestPooled \hspace{0.5pt} leads to an important 
convergence boost. MODEBI-S2 obtains the feasible solution after less than $4,000$ simulations. The rest of the 
simulation budget (until $20,000$ simulations are reached) is employed to optimize the provided solution, equivalent to 
increasing the value of the population HV (dashed red line). The six objectives are maximized or minimized (based on the 
datasheet) and diverse solution are obtained. 

We can conclude that once again the second version of the method, MODEBI-S2, acquires the best HV of the population 
(HV$\sim0.1$), although MODEBI-S1 has at its disposal a similar number of remaining simulations from the $20,000$ 
available budget for the optimization (red and blue dashed lines). The experiments conducted on the two circuits under 
test proves that MODEBI-S2 represents the leading version of the proposed algorithm for the circuit sizing task. It must be pointed out that all MODEBI scenarios require a random population of 100 solutions for each PVT corner, 
at the start of each experiment. Therefore, the proposed algorithm was evaluated on initial simulations consisting of 
$800$ simulations for the first circuit and $1,000$ simulations for the second circuit. 

\subsection{The Most Promising Algorithms Comparison - GDE3 vs MACE vs MODEBI}\label{sec:results:comparison}
During subsections \ref{sec:results:baselineGDE3} and \ref{sec:results:BO} we identified the most promising 
and advanced algorithms presented in the literature for the circuit sizing task - GDE3 for EAs and MACE for surrogate 
models algorithms, while in subsection \ref{sec:results:MODEBI} we evaluated the proposed algorithm - MODEBI - under 
different scenarios.

\begin{figure*}[h]
    \centering
    \includegraphics[width=1\textwidth,clip]{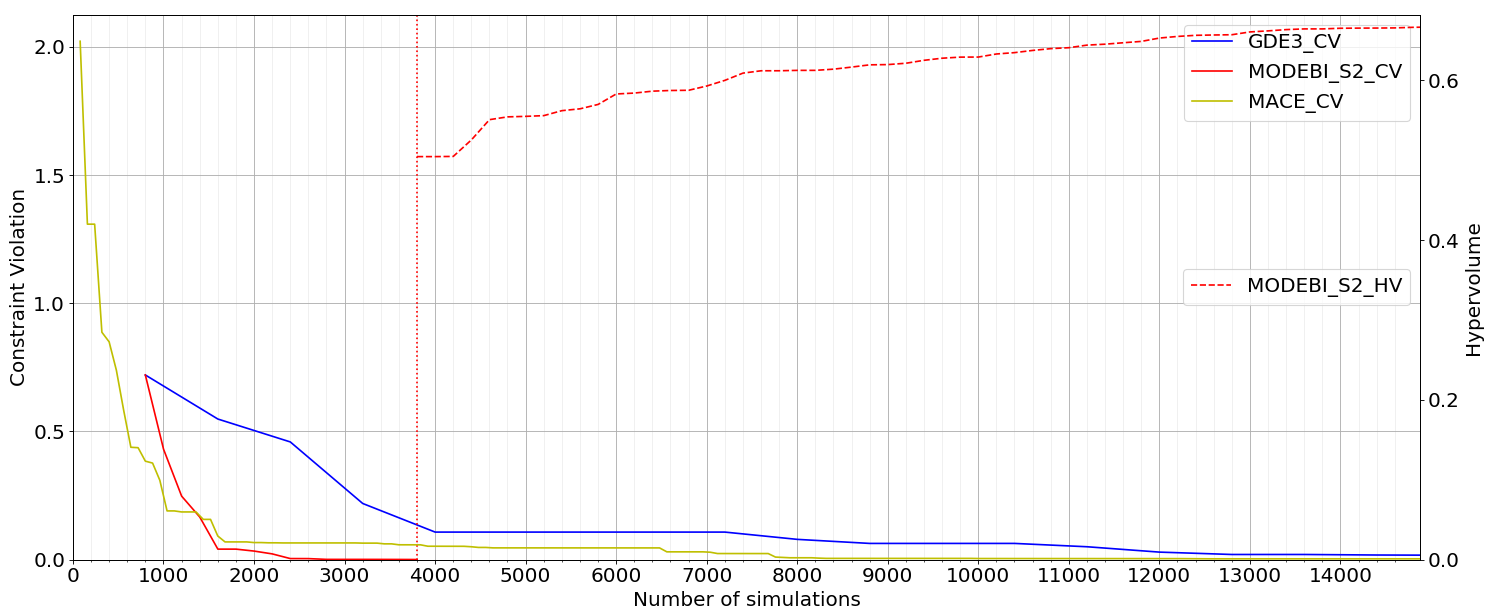}  
    \caption{The evolution of the CV for the best solution in the population (solid line) 
	and the evolution of the population HV (dashed line) for the population with feasible solutions for the
    GDE3, MACE and MODEBI-S2 on circuit 1}
    \label{fig:GDE3MACEMODEBIB1}
\end{figure*}

Therefore, the obvious next step of our analysis is a comparison between the best version of MODEBI and GDE3 and MACE.
Figure \ref{fig:GDE3MACEMODEBIB1} illustrates the experimental results obtained when applying the three algorithms on the 
first circuit (three objectives to be optimized). First, we employed MODEBI-S2, since 
the combination between \newSurvival \hspace{0.5pt} survival policy and \bestPooled \hspace{0.5pt} for offspring 
selection represents the leading variant of the method that we introduced. 

Once again, it is important to emphasize that evolutionary algorithms (i.e. GDE3 and MODEBI) generate a random population of 100 solutions 
at the start of the experiment, for each of the operating conditions corners. MACE acts like an exception to this, 
since it does not require an initial population of 100 solutions, but only five solutions for the first GP's training. As a result, 
MACE (solid yellow line) converges faster during the first $1,000$ simulations, to a CV$\sim0.2$, while GDE3 and MODEBI-S2 are still busy generating 
the initial population. Nonetheless, MACE is not able to find any feasible solution given the $15,000$ simulations budget.

GDE3 (solid blue line) displays a slower convergence; the continuous descending trend of the CV demonstrates that the best solution becomes
better as the population evolves. Still, for circuit 1 this decrease is slow and automatically leads to a full use up of 
the simulations budget without obtaining any feasible solution.

Our proposed algorithm, MODEBI-S2, represents the only algorithm capable of finding feasible solutions. 
The boost in convergence speed is impressive, since MODEBI-S2 achieves the first solution that meets all the 
constraints after $3,800$ simulations. Consequently, from this point further, the rest of the simulations budget is spend on 
optimizing the solutions, under the form of maximization of the three objectives and increasing the diversity of the population. 
This optimization translates into a HV's improvement, reaching a maximum of almost HV$\sim0.7$ (dashed red line). 

\begin{figure*}[h]
    \centering
    \includegraphics[width=1\textwidth,clip]{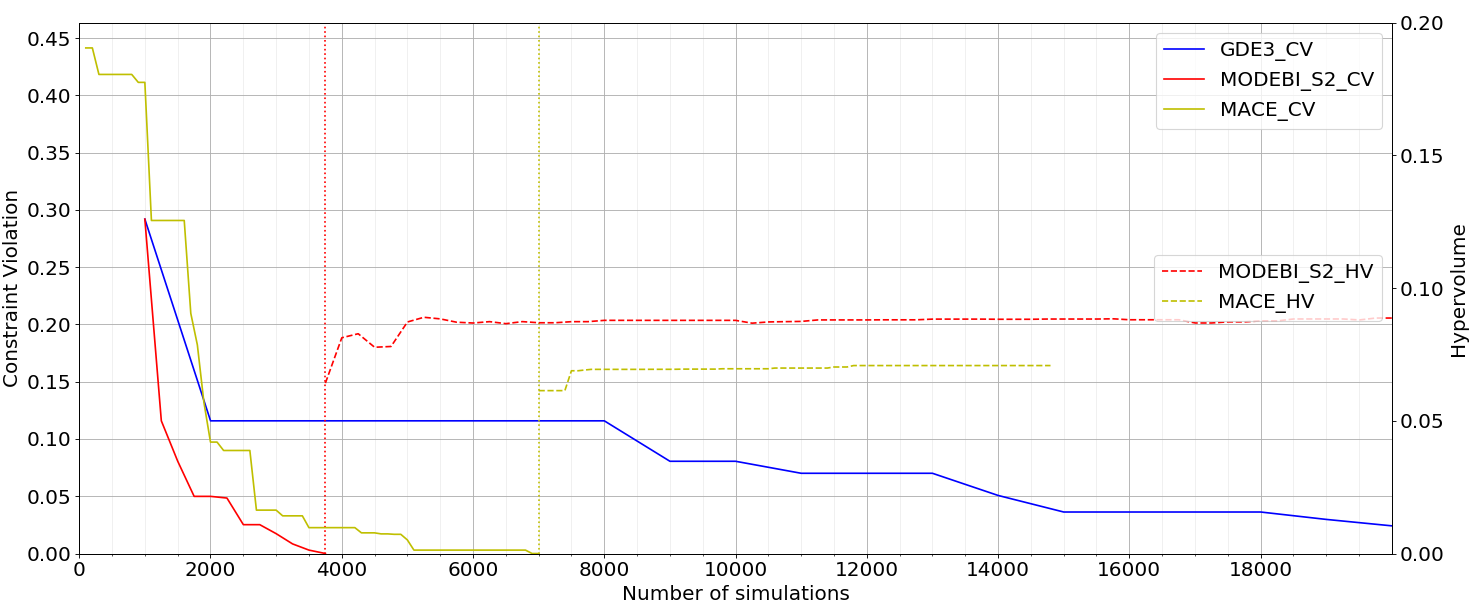}  
    \caption{The evolution of the CV for the best solution in the population (solid line) 
	and the evolution of the population HV (dashed line) for the population with feasible solutions for the
    GDE3, MACE and MODEBI-S2 on circuit 2}
    \label{fig:GDE3MACEMODEBIL1}
\end{figure*}

When applying the algorithms on the second circuit under test, comprising six objectives, we obtain the results represented  
in Figure \ref{fig:GDE3MACEMODEBIL1}. In a similar way to the results obtained on circuit 1, GDE3 (solid blue line) 
is characterized by the slowest convergence, reaching the best CV (CV$\sim0.025$) at the end of the simulation budget. 
Therefore, this algorithm is not capable of returning any feasible solutions.

On the other hand, both MACE (solid yellow line) and our proposed algorithm, MODEBI-S2 (solid red line) managed to achieve 
feasible solutions for the six objectives to be optimized. By comparing the two algorithms, we can easily identify that 
MODEBI-S2 converges rapidly and reaches a population consisting in only solutions that meet all the constraints after 
$3,700$ simulations. At the same time, MACE displays a fair convergence speed, to a CV$\sim0.05$ after less than $3,000$ 
simulations, spending a total number of $7,000$ simulations to reach the first feasible solution. 

Out of these points, both algorithms deal with the solution optimization until the entire simulation budget ($20,000$ 
simulations) is used up (dashed yellow and red lines). The HV's improvement for MODEBI-S2 is excelling, spanning to a 
HV$\sim0.09$, compared to a HV$\sim0.07$ obtained by MACE. 

In summary, the association between \bestPooled \hspace{0.5pt} for offspring selection and \newSurvival \hspace{0.5pt} 
survival policy introduced by MODEBI-S2 leads to the best performance of the three considered algorithms in all 
considered experiments. Our proposed algorithm represents the most promising method displaying flexibility when a large 
number of objectives is involved, besides a great capability to identify diversified solutions in what regards the circuit 
sizing task.  

\paragraph{Required Time}
A very important aspect when it comes to circuit sizing and optimization is the actual time (minutes) required to perform this task, in order to offer the decision makers an accurate estimation. 
The comparison presented in the previous subsections was performed in terms of number of simulations and it represents a very 
important aspect, but it is not complete because it only offers a glimpse on how quickly the best or even the feasible 
solutions can be reached. 

Practically, until now we considered that the simulator used for the circuits does not allow parallelism when it comes 
to simulations. But factoring in that the proprietary-simulator employed for the experiments can run simulations in 
parallel and since GDE3 and MODEBI-S2 propose more than one candidate to be evaluated at each iteration, we decided to complete the analysis 
with a comparison over time of the considered algorithms. The limit was set to $50$ simulations in parallel, as the 
simulator allows.

\begin{figure*}[h]
    \centering
    \includegraphics[width=1\textwidth,clip]{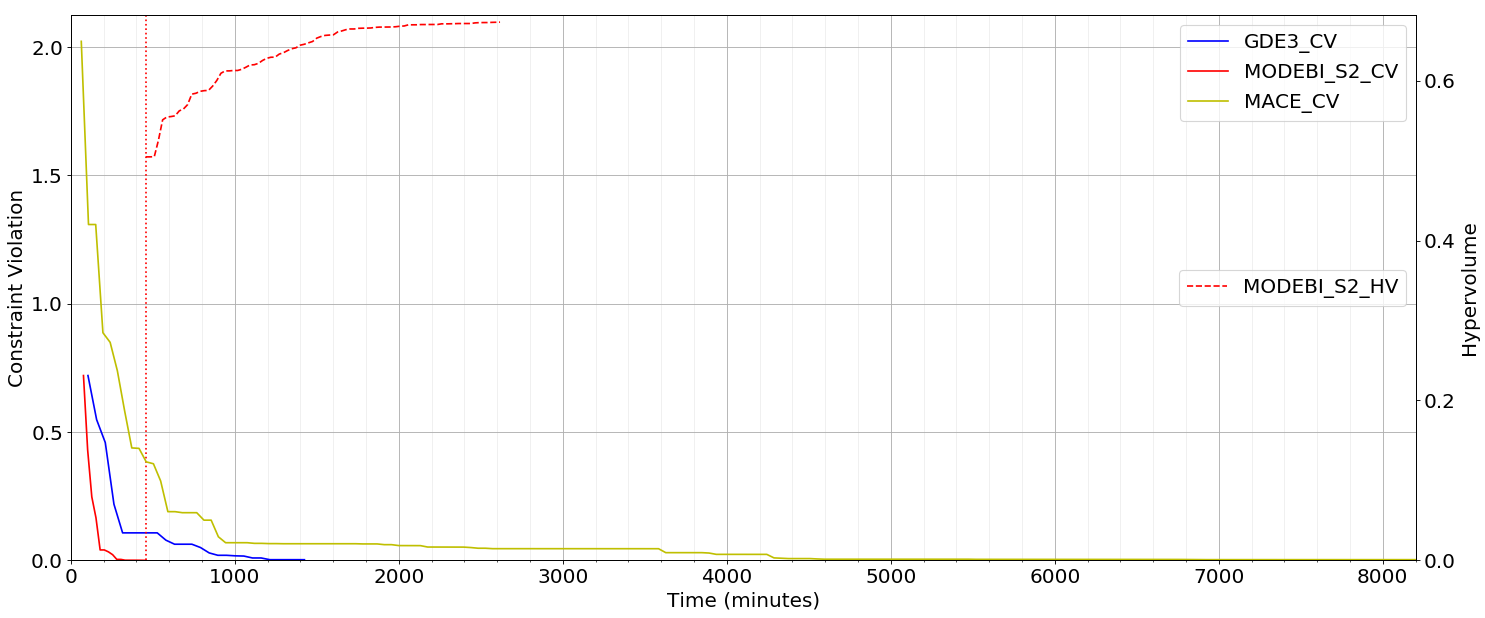}  
    \caption{The evolution in time of the CV for the best solution in the population (solid line) 
	and the evolution of the population HV (dashed line) for the population with feasible solutions for the
    GDE3, MACE and MODEBI-S2 on circuit 1}
    \label{fig:GDE3MACEMODEBIB1timing}
\end{figure*}

In Figure \ref{fig:GDE3MACEMODEBIB1timing}, we compared the three algorithms in terms of the necessary minutes to achieve 
either the best possible solution, or the feasible solutions followed by the HV improvement for the first circuit. 
Correlated to the experimental results presented in Figure \ref{fig:GDE3MACEMODEBIB1}, we easily observe that our proposal 
represents the most time-efficient algorithm. MODEBI-S2 reaches feasible solutions after approximately $400$ minutes, 
and it requires $2,200$ more minutes to reach the best HV of the population. 
While GDE3 and MACE were not able to identify feasible solutions, the proposed method needed only $2,600$ minutes to 
achieve the maximized objectives (all three), through a varied population of solutions.

\begin{figure*}[h]
    \centering
    \includegraphics[width=1\textwidth,clip]{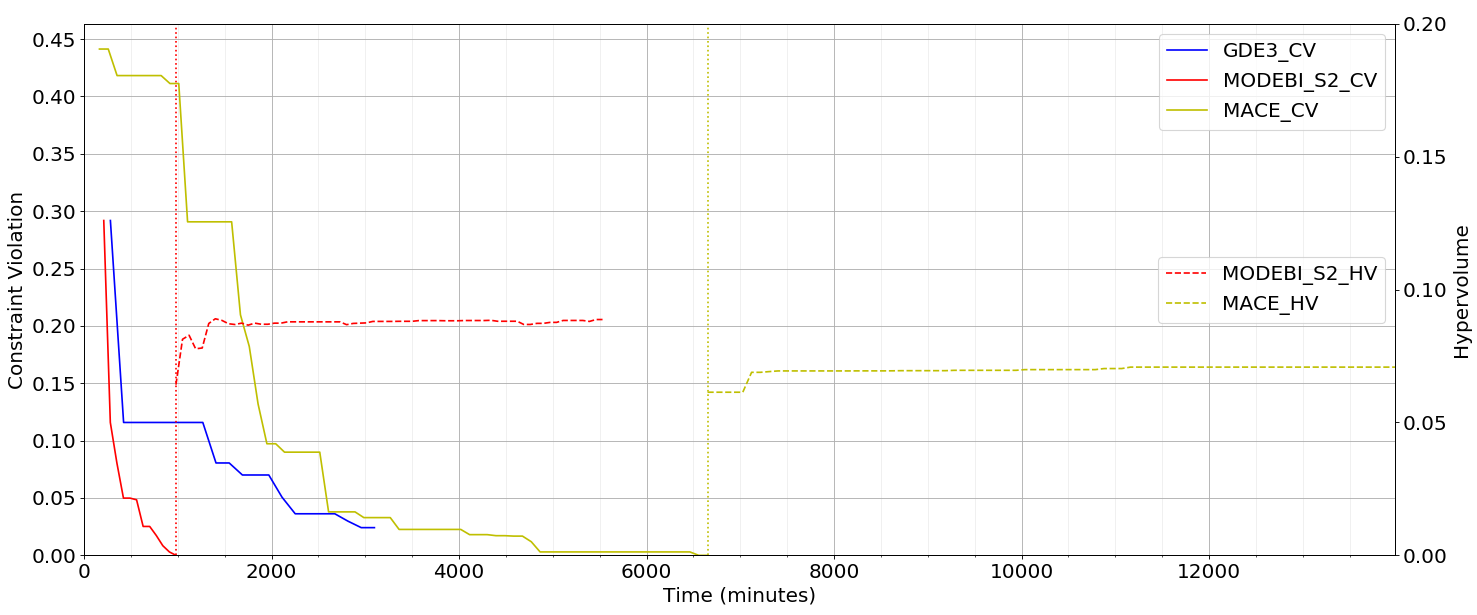}  
    \caption{The evolution in time of the CV for the best solution in the population (solid line) 
	and the evolution of the population HV (dashed line) for the population with feasible solutions for the
    GDE3, MACE and MODEBI-S2 on circuit 2}
    \label{fig:GDE3MACEMODEBIL1timing}
\end{figure*}

Figure \ref{fig:GDE3MACEMODEBIL1timing} illustrates the results of the same experiment conducted on circuit 2. 
The benefits of MODEBI-S2 are easy to be identified: it achieves the first feasible solution after $1,000$ minutes 
and it spends $5,500$ more minutes to bring in the best HV of the population (HV$\sim0.09$). Since MACE does not allow 
$50$ parallel simulations, but only uses ten simulations in parallel (all corners for one solution), 
it requires more time to reach the feasible solutions (more than $6,500$ minutes) and exhaust the rest of the 
simulation budget on improving the HV, scoring a maximum of $\sim0.07$. 

Moreover, since GDE3 and MODEBI-S2 are population-based algorithms, they exhaust the maximum number of simulations provided in 
parallel, and they have similar run times as the two figures display. In total, it can be observed that MODEBI-S2 takes 
a little longer, because training the GP and selecting the population when there are many feasible solutions takes a 
significant amount of time. Regardless, it remains the only algorithm able to identify feasible solutions for both 
circuits under test. 



\begin{figure*}[h]
    \centering
    \includegraphics[width=0.75\textwidth,trim={0 0 0
    2pt},clip]{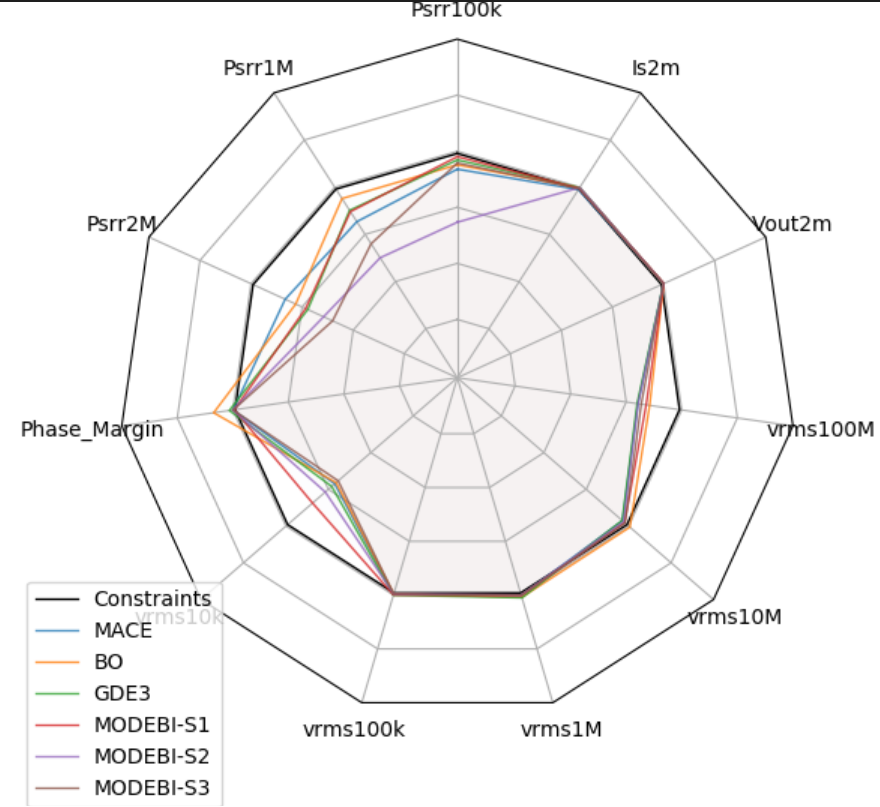}  
    \caption{The best solutions generated by MACE, BO, GDE3, MODEBI-S1,
    MODEBI-S2, MODEBI-S3 for the three objectives of circuit 1}
    \label{fig:GDE3MACEMODEBIspider}
\end{figure*}

\paragraph{Best Solution Comparison}
So far, the current section compares the various optimization algorithms in
terms of aggregate score metrics such as the lowest CV in the current population
and the population HV. For a more practical comparison, we next inspect the best
solution obtained by each of the six algorithms in terms of circuit responses.
As a note, the HV metric can be computed on a non-empty set of solutions. In
this section we have computed the HV using multiple solutions when comparing the
algorithms in the previous figures. For Figure \ref{fig:GDE3MACEMODEBIspider},
we aim to show the best solution found by each algorithm, and thus we compute
the HV using one solution at a time and select the one with the largest
individual HV. 
The HV is computed using the 3 objectives of the first circuit. It is worth mentioning
that GDE3, BO, MACE and MODEBI-S1 were not able to identify feasible solutions,
while the other two versions of MODEBI (MODEBI-S2 and MODEBI-S3) did.

The 11 axes of the spider chart represent the responses. The solid black line
polygon marks the constraints of this circuit. The colored lines correspond to
the responses of the best solutions for each algorithm, chosen as described in
the previous paragraph. A feasible solution will fall inside the black
constraints polygon -- the closer to the center, the better.
MACE, BO, GDE3 and MODEBI-S1 manage to obtain solutions with good values for the
three objectives (\ie the PSRRs), but fail shortly to meet some constraints (\eg
Phase Margin). The solution found by MODEBI-S2 is clearly the best solution
illustrated in the spider chart. Since it reached feasible solutions faster, the
algorithm had ample leftover budget to explore the hyperspace in detail
generating, in the end, many solutions in the optimal regions.
MODEBI-S3 is the other algorithm to reach feasible solutions for this circuit.
The best solution found is non-dominated when compared to the solution found by
MODEBI-S2: it has a slightly better PSRR2M, a worse PSRR1M and a significantly
worse PSRR100k. As such, it is probably a lower quality compromise between the
three objectives. In any event, a designer should be presented with all
non-dominated solutions found (the Pareto front) by any algorithm or
combinations of algorithms so that he can select the best trade-off between the
objectives. 

\subsection{Randomness Impact}\label{sec:results:randomness}
Since all algorithms presented in this section require a randomly sampled
initial population, regardless of the category they belong to, this introduces a
certain degree of randomness in the optimization. Even more, evolutionary
algorithms are stochastic by nature since the offspring generation relies on
random crossover and mutation operations. 

In this subsection we investigate how large is the impact of the randomness on
the optimization results and how robust the algorithms are to starting from new
random initial populations. Each algorithm presented in this section was run
three times, each run using one of three starting populations generated randomly
from different random seeds - \emph{seed1}, \emph{seed2}, \emph{seed3}. All the
results presented in this section so far were obtained when \emph{seed1} was
employed; to quantify the randomness impact, we will present a comparison
between \emph{seed1}, \emph{seed2} and \emph{seed3} on circuit 1. The randomness
impact on the two circuits is similar, and so only one circuit is presented to
simplify the discussion.

\begin{figure}[!h]
    \centering
    \begin{subfigure}[b]{1\textwidth}
       \includegraphics[width=1\linewidth]{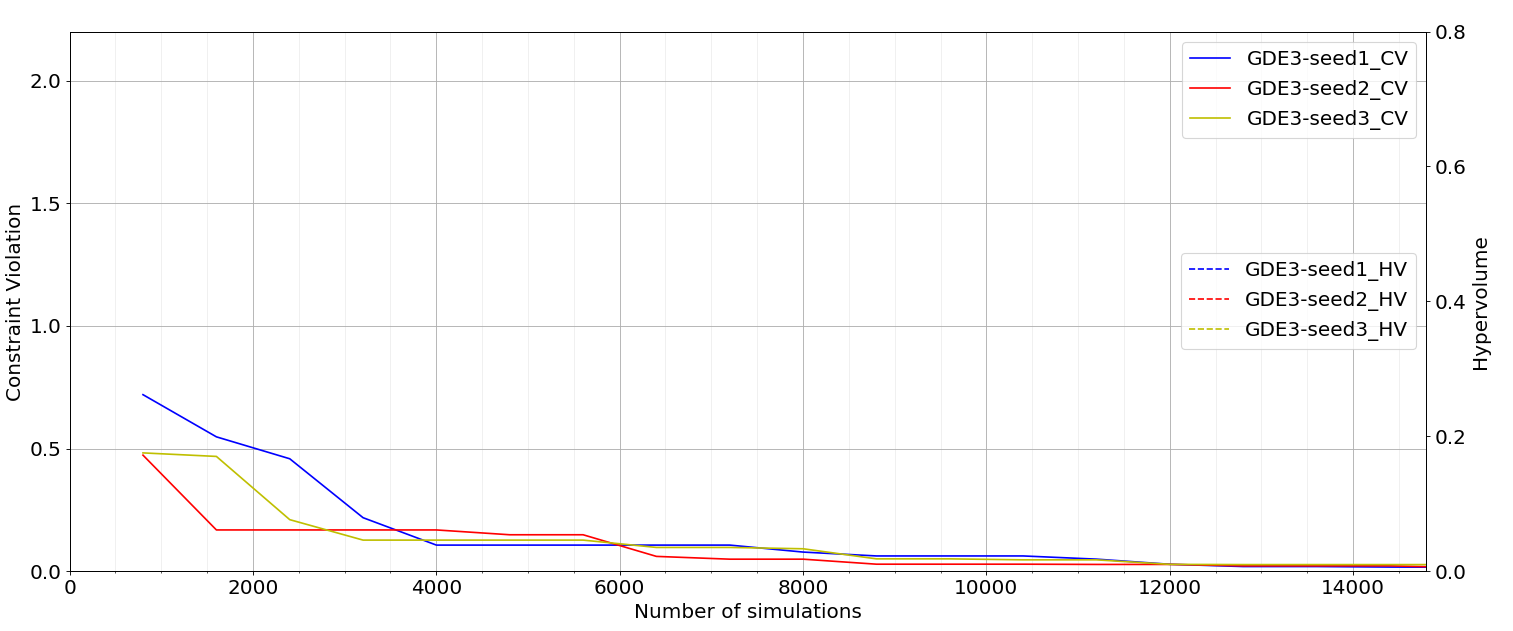}
       \caption{}
       \label{fig:Ng1} 
    \end{subfigure}
    
    \begin{subfigure}[b]{1\textwidth}
       \includegraphics[width=1\linewidth]{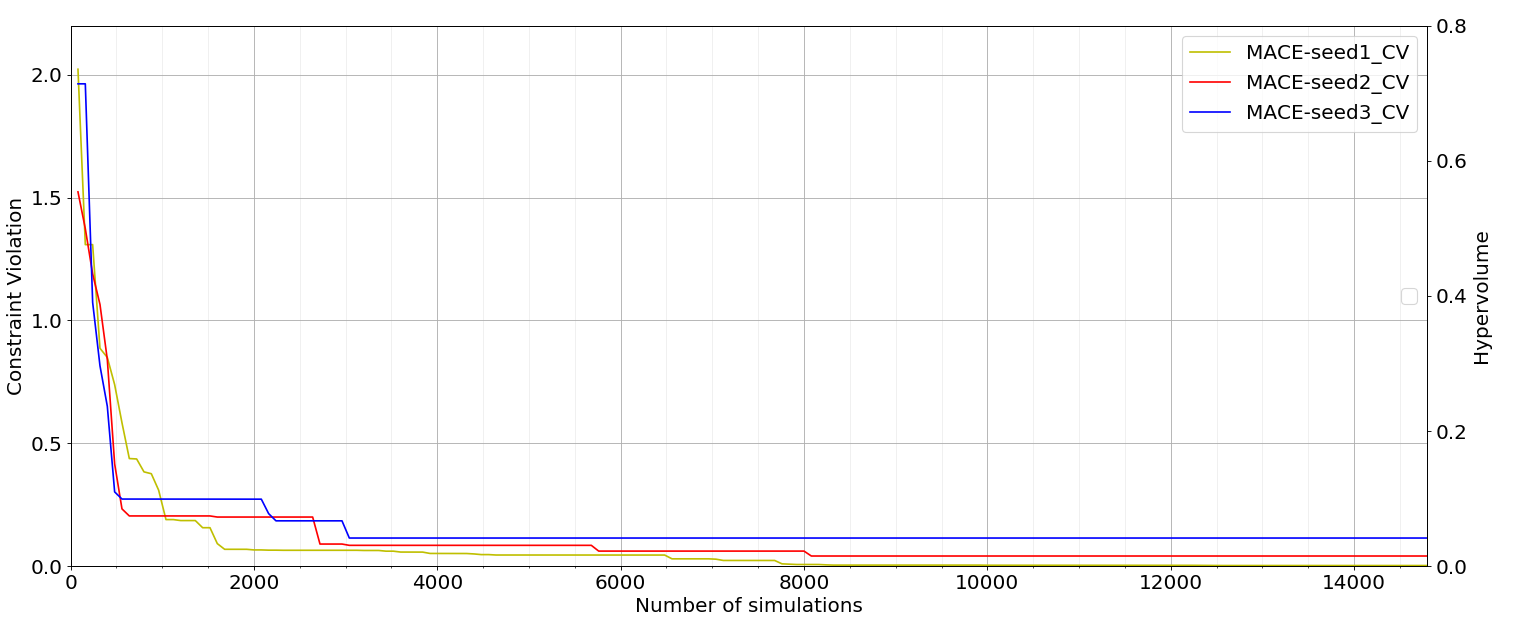}
       \caption{}
       \label{fig:Ng2}
    \end{subfigure}
    
    \begin{subfigure}[b]{1\textwidth}
        \includegraphics[width=1\linewidth]{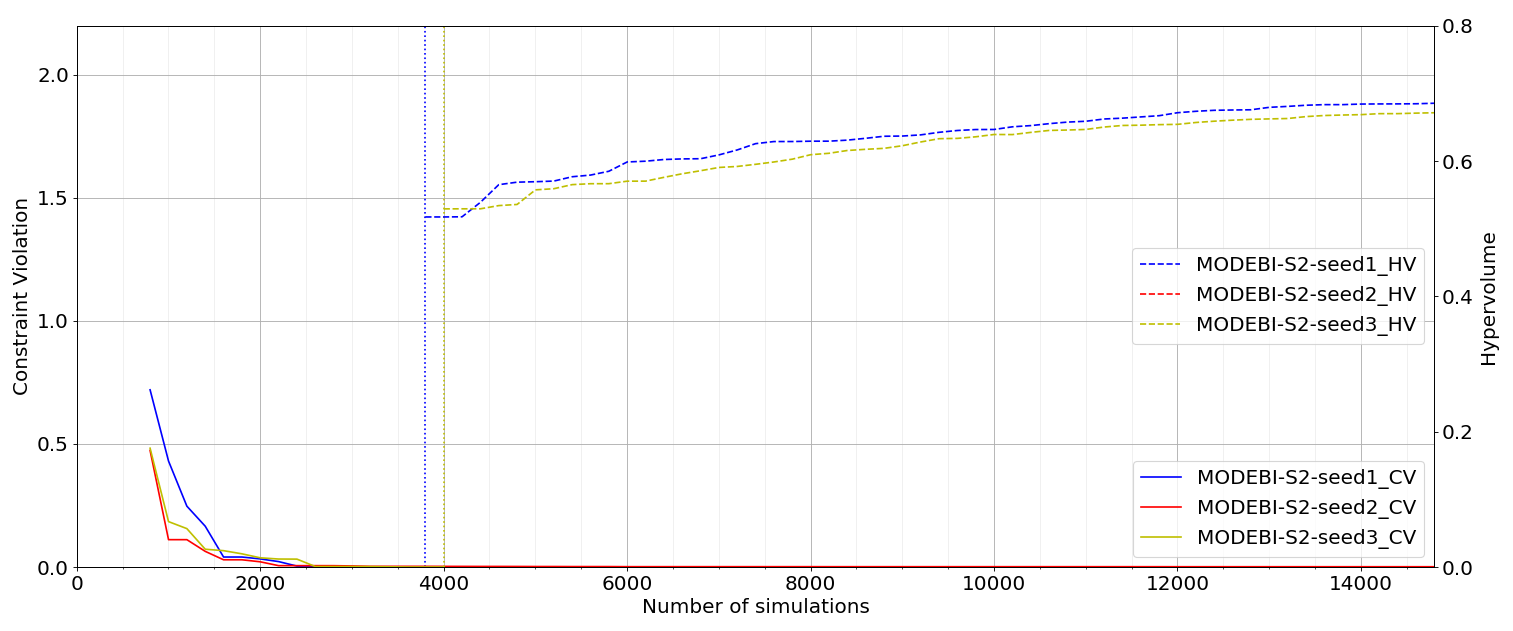}
        \caption{}
        \label{fig:Ng3}
    \end{subfigure}

    \caption{The evolution of the CV for the best solution in the population (solid line) 
    and the evolution of the population HV (dashed line) with a $15,000$ simulations 
    budget on circuit 1 using 3 seeds for a) GDE3 b) MACE c) MODEBI-S2}
    \label{fig:GDE3vsMACEvsMODEBI3seeds} 
\end{figure}

Figure \ref{fig:GDE3vsMACEvsMODEBI3seeds} shows the difference between the 3
seeds for the following algorithms: GDE3, MACE and MODEBI-S2. BO was not included 
since MACE has better performance. Similarly, only MODEBI-S2 is presented since the 
other two MODEBI scenarios have inferior performance. 

The results confirm that randomness has an important impact on the performance. 
The most affected algorithm was MACE (Figure
\ref{fig:GDE3vsMACEvsMODEBI3seeds} (b)). Two of the runs (\emph{seed2} and
\emph{seed3}) have slower convergence compared to the run with \emph{seed1}.
Moreover, after $8000$ simulations only the run with \emph{seed1} is still able
to improve solution quality. Starting in a certain area of the hyperspace has a large
impact on MACE's optimization. 

The algorithms based on evolutionary computation are slightly less affected by
randomness, compared to MACE. However, even though after $2500$ simulations it
is hard to see any difference between the three MODEBI-S2 runs, the run with
\emph{seed2} is unable to find any feasible solution while the other two find
them after roughly $4000$ simulations. 

In conclusion, randomness can have a significant impact on the results. While 
it is not enough to change the ranking between the algorithms, it can lead to 
the occasional bad run such as MODEBI-S2 with \emph{seed2}. To limit this impact
and increase the algorithms' robustness, one should consider potential
counter-measures, among which we mention: ensuring a better
distribution of the initial population instead of simple uniform sampling,
using restarts or even enabling a more aggressive exploration which might mitigate
the randomness impact by ensuring that more areas of the decision space are
visited. These methods will be explored in future work.

\afterpage{\FloatBarrier}

\section{Conclusions and Future Work}\label{sec:conclusions}

In this paper, we proposed an innovative multi-objective optimization algorithm
for automated circuit sizing - MODEBI. Our contribution is designed to deal with circuits
with a high number of design variables (order of tens), several PVT corners and many (10+)
conflicting responses that need to meet the specifications. In this context, MODEBI represents a 
design optimization method that uses an EA inspired from GDE3 to explore 
the complex hyperspace of the design variables and PVT corners. In addition, our solution uses GPs as a
surrogate for expensive circuit simulations, effectively boosting the convergence of the
EA.

Currently, most state-of-the-art algorithms solve the multi-objective tasks by trimming them to 
single-objective optimization or even including a priori bias. We address these by employing the Pareto dominance; 
based on it, we explore directly over the multi-objective space. This way, we are able to guarantee a population 
of various solutions to circuit designers. 

The paper explored two survival policies and we showed that our proposal (\newSurvival)
leads to finding feasible solutions using the allocated simulations budget. In contrast,
GDE3 -- both with or without the addition of the GP surrogate model -- with conventional
survival and selection mechanisms could not find solutions that satisfy the constraints.
Furthermore, when combining \newSurvival \hspace{0.5pt} survival with the proposed
offspring selection method (\bestPooled), our method shows an even faster convergence and
finds better final solutions. All the experiments were performed on a voltage regulator.

Our future work includes exploiting the potential correlations between circuit responses to
further increase the convergence speed, as well as exploring an integration of Deep Neural
Networks, which could be used to reduce the input space by mapping the original data to a
lower dimensional feature space on which GPs can then be applied. As presented in subsection 
\ref{sec:results:randomness}, reducing the randomness impact and increasing the proposed 
algorithm robustness will be the subject of forthcoming research.

Finally, another interesting avenue of research is augmenting the algorithm with
individual search heuristics capable of performing local refinements, in a similar way to
memetic algorithms. These local search processes can take advantage of a differentiable
surrogate model to solve the inverse problem via gradient search. Thus, the use of
acquisition functions could potentially be integrated in such a way that proposing
multiple diverse candidates at each iteration is practicable and even faster convergence
speed is achieved.

\section*{Acknowledgement}
This work was supported by a grant of the Romanian Ministry of Education and Research, CCCDI - UEFISCDI, project number PN-III-P2-2.1-PTE-2019-0861, within PNCDI III.


\bibliography{references.bib}

\end{document}